\documentclass[12pt]{article}
\usepackage{amsmath}
\usepackage{amsfonts}
\usepackage{amssymb}

\usepackage{times}
\usepackage{graphicx}
\usepackage{color}
\usepackage{multirow}

\usepackage{natbib}

\usepackage{rotating}
\usepackage{bbm}
\usepackage{latexsym}

\usepackage{url}
\usepackage{authblk}
\usepackage[labelformat=simple]{subcaption} 
\newcommand\independent{\protect\mathpalette{\protect\independenT}{\perp}}
\def\independenT#1#2{\mathrel{\rlap{$#1#2$}\mkern2mu{#1#2}}}

\newcommand{\argmin}{\mathop{\mathrm{argmin}}}
\newcommand{\argmax}{\mathop{\mathrm{argmax}}}

\newcommand{\bx}{\boldsymbol{x}}
\newcommand{\by}{\boldsymbol{y}}
\newcommand{\bz}{\boldsymbol{z}}
\newcommand{\bX}{\boldsymbol{X}}
\newcommand{\bY}{\boldsymbol{Y}}
\newcommand{\bZ}{\boldsymbol{Z}}

\newcommand{\dx}{\mathrm{d}\boldsymbol{x}}
\newcommand{\dy}{\mathrm{d}\boldsymbol{y}}
\newcommand{\dz}{\mathrm{d}\boldsymbol{z}}

\newcommand{\bv}{\boldsymbol{v}}
\newcommand{\bu}{\boldsymbol{u}}
\newcommand{\bW}{\boldsymbol{W}}

\newcommand{\btheta}{\boldsymbol{\theta}}
\newcommand{\bhtheta}{\widehat{\btheta}}
\newcommand{\balpha}{\boldsymbol{\alpha}}
\newcommand{\bhalpha}{\widehat{\balpha}}
\newcommand{\bw}{\boldsymbol{w}}

\newcommand{\bphi}{\boldsymbol{\phi}}
\newcommand{\bvphi}{\boldsymbol{\varphi}}

\newcommand{\bpsi}{\boldsymbol{\psi}}

\newcommand{\bh}{\boldsymbol{h}}

\newcommand{\bhh}{\widehat{\boldsymbol{h}}}

\newcommand{\bH}{\boldsymbol{H}}

\begin{document} 

\title{
Direct Estimation of the Derivative of Quadratic Mutual Information
with Application in Supervised Dimension Reduction
}

\author{
Voot Tangkaratt, Hiroaki Sasaki, and Masashi Sugiyama
}

\affil{The University of Tokyo}

\date{}





\maketitle

\sloppy

\begin{abstract}
A typical goal of supervised dimension reduction is to find 
a low-dimensional subspace of the input space 
such that the projected input variables preserve 
maximal information about the output variables.
The dependence maximization approach
solves the supervised dimension reduction problem 
through maximizing a statistical dependence
between projected input variables and output variables.
A well-known statistical dependence measure is  
mutual information (MI) which is based on the Kullback-Leibler (KL) divergence.
However, it is known that the KL divergence is sensitive to outliers. 
On the other hand, quadratic MI (QMI) is a variant of MI based on the 
$L_2$ distance which is more robust against outliers than the KL divergence,
and a computationally efficient method to estimate QMI from data, 
called least-squares QMI (LSQMI), has been proposed recently.
For these reasons, developing a supervised dimension reduction method
 based on LSQMI seems promising.
However, not QMI itself, 
but the derivative of QMI is needed for subspace search in supervised dimension reduction,
and the derivative of an accurate QMI estimator 
is not necessarily a good estimator of the derivative of QMI.
In this paper, we propose to directly estimate the derivative of QMI
without estimating QMI itself.
We show that the direct estimation of the derivative of QMI
is more accurate than the derivative of the estimated QMI.
Finally, we develop a supervised dimension reduction
algorithm which efficiently uses the proposed derivative estimator,
and demonstrate through experiments that the proposed method 
is more robust against outliers than existing methods.
\end{abstract} 

\section{Introduction}

Supervised learning is one of the central problems in machine learning
which aims at learning an input-output 
relationship from given input-output paired data samples.
Although many methods were proposed to perform supervised learning,
they often work poorly when 
the input variables have high dimensionality.
Such a situation is commonly referred to as 
the \textit{curse of dimensionality} \citep{Bishop:2006:PRM:1162264},
and a common approach to mitigate the curse of dimensionality is 
to preprocess the input variables by \textit{dimension reduction}
\citep{DBLP:journals/ftml/Burges10}.

A typical goal of dimension reduction in supervised learning is
to find a low-dimensional subspace of the input space
such that the projected input variables preserve 
maximal information about the output variables.
Thus, a successive supervised learning method can 
use the low-dimensional projection of the input variables
to learn the input-output relationship 
with a minimal loss of information. 
The purpose of this paper is to develop a 
novel supervised dimension reduction method.

The dependence maximization approach solves 
the supervised dimension reduction problem
through maximizing a statistical dependence measure 
between projected input variables and output variables.
\textit{Mutual information} (MI) is a well-known tool
for measuring statistical dependency between random variables 
\citep{Cover:1991:EIT:129837}.
MI is well-studied 
and many methods were proposed to estimate MI from data.
A notable method 
is the \textit{maximum likelihood MI} (MLMI)
\citep{DBLP:journals/jmlr/SuzukiSSK08},
which does not require any assumption on the data distribution
and can perform model selection via cross-validation.
For these reasons, MLMI seems to be an appealing tool 
for supervised dimension reduction.
However, MI is defined based on the
\textit{Kullback-Leibler} divergence \citep{Kullback51klDivergence},
which is known to be sensitive to outliers \citep{Basu1998}.
Hence, MI is not an appropriate tool when 
it is applied on a dataset containing outliers.

\textit{Quadratic MI} (QMI) 
is a variant of MI \citep{DBLP:journals/vlsisp/PrincipeXZF00}.
Unlike MI, QMI is defined based on the $L_2$ distance.
A notable advantage of the $L_2$ distance
over the KL divergence is that the $L_2$ distance
is more robust against outliers \citep{Basu1998}.
Moreover, a computationally efficient method to estimate QMI from data, 
called \textit{least-squares QMI} (LSQMI) 
\citep{DBLP:journals/ieicet/SainuiS13}, was proposed recently.
LSQMI does not require any assumption on the data distribution
and it can perform model selection via cross-validation.
For these reasons,
developing a supervised dimension reduction method 
based on LSQMI is more promising.

An approach to use LSQMI for supervised dimension reduction 
is to firstly estimate QMI between projected input variables 
and output variables by LSQMI, 
and then search for a subspace which maximizes the estimated QMI 
by a nonlinear optimization method such as gradient ascent.
However, the essential quantity of 
the subspace search is the derivative of QMI 
w.r.t.~the subspace, not QMI itself.
Thus, LSQMI may not be an appropriate tool
for developing supervised dimension reduction methods
since the derivative of an accurate QMI estimator is not
necessarily an accurate estimator of the derivative of QMI. 

To cope with the above problem,
in this paper, we propose a novel method to \textit{directly} 
estimate the derivative of QMI without estimating QMI itself.
The proposed method has the following advantageous properties:
it does not require any assumption on the data distribution,
the estimator can be computed analytically,
and the tuning parameters can be objectively chosen by cross-validation.
We show through experiments that 
the proposed direct estimator of the derivative of QMI
is more accurate than the derivative of the estimated QMI.
Then we develop a fixed-point iteration 
which efficiently uses the proposed estimator of the derivative of QMI
to perform supervised dimension reduction.
Finally, we demonstrate the usefulness of the 
proposed supervised dimension reduction method 
through experiments and show that 
the proposed method is more robust against outliers 
than existing methods.

The organization of this paper is as follows.
We firstly formulate the supervised dimension reduction problem 
and review some of existing methods in Section~\ref{section:SDR},
Then we give an overview of QMI and 
review some of QMI estimators in Section~\ref{section:QMI}.
The details of the proposed derivative estimator are given in Section~\ref{section:LSQMID}.
Then in Section~\ref{section:SDR_LSQMID} 
we develop a supervised dimension reduction algorithm
based on the proposed derivative estimator.
The experiment results are given in Section~\ref{section:experiments}.
The conclusion of this paper is given in Section~\ref{section:conclusion}.

\section{Supervised Dimension Reduction}
\label{section:SDR}
In this section, we firstly formulate the supervised dimension reduction problem.
Then we briefly review existing supervised dimension reduction methods and discuss their problems.

\subsection{Problem Formulation}
Let $\mathcal{D}_{\boldsymbol{\mathrm{x}}} \subset \mathbb{R}^{d_{\boldsymbol{\mathrm{x}}}}$ and 
$\mathcal{D}_{\boldsymbol{\mathrm{y}}} \subset \mathbb{R}^{d_{\boldsymbol{\mathrm{y}}}}$ be 
the input domain and output domain 
with dimensionality $d_{\boldsymbol{\mathrm{x}}}$ and 
$d_{\boldsymbol{\mathrm{y}}}$, respectively,
and $p(\bx,\by)$ be a joint probability density on
$\mathcal{D}_{\boldsymbol{\mathrm{x}}} \times \mathcal{D}_{\boldsymbol{\mathrm{y}}}$.
Firstly, assume that we are given an input-output paired data set 
$\mathcal{D} = \{(\bx_i, \by_i)\}_{i=1}^n$,
where each data sample is drawn independently from the joint density:
\begin{align*}
\{(\bx_i, \by_i)\}_{i=1}^n \stackrel{i.i.d.}{\sim} p(\bx,\by).
\end{align*}

Next, let $\bW \in \{\bW \in \mathbb{R}^{d_{\boldsymbol{\mathrm{z}}} \times d_{\boldsymbol{\mathrm{x}}}}| \bW\bW^\top = \boldsymbol{I}_{d_{\boldsymbol{\mathrm{z}}}}\}$ 
be an orthonormal matrix 
with a known constant $d_{\boldsymbol{\mathrm{z}}} \leq d_{\boldsymbol{\mathrm{x}}}$,
where $\boldsymbol{I}_{d_{\boldsymbol{\mathrm{z}}}}$ denotes the
$d_{\boldsymbol{\mathrm{z}}}$-by-$d_{\boldsymbol{\mathrm{z}}}$
identity matrix  and $^\top$ denotes the matrix transpose.
Then assume that there exists a $d_{\boldsymbol{\mathrm{z}}}$-dimensional subspace in $\mathbb{R}^{d_{\boldsymbol{\mathrm{x}}}}$
spanned by the rows of $\bW$ such that 
the projection of $\bx$ onto this subspace denoted by~$\bz = \bW\bx$~
preserves the maximal information about $\by$ of $\bx$.
That is, we can substitute $\bx$ by $\bz$
with a minimal loss of information about $\by$.
We refer to the problem of estimating $\bW$ from the given data
as \textit{supervised dimension reduction}.
Below, we review some of the existing supervised dimension reduction methods.

\subsection{Sliced Inverse Regression}

Sliced inverse regression (SIR) \citep{Li:1991} 
is a well known supervised dimension reduction method.
SIR formulates supervised dimension reduction as 
a problem of finding $\bW$ which makes 
$\bx$ and $y$ conditionally independent given $\bz$:
\begin{align}
( \bx \independent y )~|~\bz. \label{SDR_condition_ind}
\end{align}

The key principal of SIR lies on the following equality
\footnote{For simplicity, we assume that $\bx$ is standardized 
so that $\mathbb{E}[\bx] = 0$ and 
$\mathbb{E}[\bx\bx^\top] = \boldsymbol{I}_{d_{\boldsymbol{\mathrm{z}}}}$}
\begin{align}
\mathbb{E}[\boldsymbol{c}^\top\bx | \bW\bx]
&= a_0 + \sum_{i=1}^{d_{\boldsymbol{\mathrm{z}}}} 
   a_i \boldsymbol{w}_i^\top \bx,	\label{SIR_condition}
\end{align}
where $\mathbb{E}$ denotes the conditional expectation 
and $\bw_i$ denotes the $i$-th row of $\bW$.
The importance of this equality 
is that if the equality holds for any 
$\boldsymbol{c} \in \mathbb{R}^{d_{\boldsymbol{\mathrm{x}}}}$
and some constants 
$a_0, a_1, \dots, a_{d_{\boldsymbol{\mathrm{z}}}}$,
then the inverse regression curve $\mathbb{E}[\bx|y]$ 
lies on the space spanned by $\bW$ which satisfies Eq.\eqref{SDR_condition_ind}.
Based on this fact, 
SIR estimates $\bW$ as follows.
First, the range of $y$ is sliced into multiple slices.
Then $\mathbb{E}[\bx|y]$ is estimated 
as the mean of $\bx$ for each slice of $y$.
Finally, $\bW$ is obtained 
as the $d_{\boldsymbol{\mathrm{z}}}$
largest principal components of the covariance matrix of the means.

The significant advantages of SIR are its simplicity and 
scalability to large datasets.
However, SIR relies on the equality in Eq.\eqref{SIR_condition}
which typically requires that $p(\bx)$ is an 
elliptically symmetric distribution such as Gaussian.
This is restrictive and thus the practical 
usefulness of SIR is limited.

\subsection{Minimum Average Variance Estimation based on the Conditional Density Functions}

The \textit{minimum average variance estimation 
based on the conditional density functions} (dMAVE) 
\citep{xia2007} is a supervised dimension method 
which does not require any assumption on the data distribution
and is more practical compared to SIR.
Briefly speaking, dMAVE aims to find a matrix $\bW$ 
which yields an accurate 
non-parametric estimation of the conditional density $p(y|\bz)$.

The essential part of dMAVE is the following model:
\begin{align*}
H_{b}(\widetilde{y} - y) 
&= m_b(\bz,y) + \varepsilon_b(y|\bz),
\end{align*}
where $H_b$ denotes a symmetric kernel function with bandwidth $b > 0$,
$m_b(\bz, y)$ denotes a conditional expectation of 
$H_{b}(\widetilde{y} - y)$ given $\bz$,
and $\varepsilon_b(y|\bz) = H_b(\widetilde{y} - y) 
- \mathbb{E}\left[H_b(\widetilde{y} - y)|\bz \right]$
with $\mathbb{E}\left[\varepsilon_b(y|\bz)\right] = 0$.
An important property of this model is that 
$m_b(\bz,y) \rightarrow p(y|\bz)$
when $b \rightarrow 0$ as $n \rightarrow \infty$.
Then, dMAVE estimates $m_b(\bz, y)$ by 
a local linear smoother \citep{FanYaoTong:1996}.
More specifically, a local linear smoother of 
$m_b(\bz_i,y_k)$ is given by
\begin{align}
m_b(\bz_i,y_k) 
&\approx {m}_b(\bz_j,y_k) 
+ \frac{{\partial m}_b(\bz_j,y_k)}{\partial \bz} (\bz_i - \bz_j) \notag \\
&= a_{jk} + \boldsymbol{b}_{jk}^\top \bW (\bx_i - \bx_j), \label{LLS}
\end{align}
where $\bz_j$ is an arbitrary point close to $\bz_i$,
and $a_{jk} \in \mathbb{R}$ 
and $\boldsymbol{b}_{jk} \in \mathbb{R}^{d_{\boldsymbol{\mathrm{z}}}}$
are parameters.
Based on this local linear smoother, 
dMAVE solves the following minimization problem:
\begin{align}
\min_{\bW, a_{jk}, \boldsymbol{b}_{jk}}
\frac{1}{n^3} \sum_{j,k=1}^n 
\rho(\bx_j, y_k)
\sum_{i=1}^n 
\left[ H_b(y_i - y_k) -  a_{jk}
- \boldsymbol{b}_{jk}^\top\bW(\bx_i - \bx_j)\right]^2 K_h(\bx_i, \bx_j), \label{dMAVE_obj}
\end{align}
where  $K_h$ is a symmetric kernel function with bandwidth $h > 0$.
The function $\rho(\bx,y)$ is a trimming function 
which is evaluated as zero when the densities of $\bx$ or $y$ are lower than some threshold.
A solution to this minimization problem 
is obtained by alternatively 
solving quadratic programming problems
for $\bW$, and $(a_{jk}, \boldsymbol{b}_{jk})$ until convergence.

The main advantage of dMAVE is that it does not require 
any assumption on the data distribution.
However, a significant disadvantage of dMAVE is that 
there is no systematic method to choose 
the kernel bandwidths and the trimming threshold.
In practice, dMAVE uses a bandwidth selection method 
based on the \textit{normal-reference} 
rule of the non-parametric conditional density estimation
\citep{Silverman86, FanYaoTong:1996},
and a fixed trimming threshold. 
Although this model selection strategy works 
reasonably well in general, 
it does not always guarantee good performance on all kind of datasets.

Another disadvantage of dMAVE is that the 
optimization problem in Eq.\eqref{dMAVE_obj} may have many local solutions.
To cope with this problem, 
dMAVE proposed to use a supervised dimension reduction method 
called the \textit{outer product of gradient based on conditional density functions} (dOPG) \citep{xia2007}
to obtain a good initial solution.
Thus, dMAVE may not perform well if dOPG fails to provide a good initial solution.

\subsection{Kernel Dimension Reduction}
Another supervised dimension reduction method which does not 
require any assumption on the data distribution
is \textit{kernel dimension reduction} (KDR) \citep{fukumizu2009}.
Unlike dMAVE which focuses on the conditional density, 
KDR aims to find a matrix $\bW$ which 
satisfies the conditional independence in Eq.\eqref{SDR_condition_ind}.
The key idea of KDR is to evaluate the conditional independence through a conditional covariance operator over reproducing kernel Hilbert spaces (RKHSs) \citep{aronszajn50reproducing}. 

Throughout this subsection, we use 
$(\mathcal{H}_{\boldsymbol{\mathrm{z}}}, k_{\boldsymbol{\mathrm{z}}})$ 
to denote an RKHS of functions on the domain $\mathcal{D}_{\boldsymbol{\mathrm{z}}}$
equipped with reproducing kernel $k_{\boldsymbol{\mathrm{z}}}$:
\begin{align*}
\left\langle f, k_{\boldsymbol{\mathrm{z}}}(\cdot, \bz) \right\rangle_{\mathcal{H}_{\boldsymbol{\mathrm{z}}}} 
&= f(\bz),
\end{align*}
for $f \in \mathcal{H}_{\boldsymbol{\mathrm{z}}}$ and $\bz \in \mathcal{D}_{\boldsymbol{\mathrm{z}}}$.
The RKHSs of functions on domains $\mathcal{D}_{\boldsymbol{\mathrm{x}}}$ 
and $\mathcal{D}_{\boldsymbol{\mathrm{y}}}$ are also defined similarly as
$(\mathcal{H}_{\boldsymbol{\mathrm{x}}}, k_{\boldsymbol{\mathrm{x}}})$  
and $(\mathcal{H}_{\boldsymbol{\mathrm{y}}}, k_{\boldsymbol{\mathrm{y}}})$, respectively.
The \textit{cross-covariance operator} $\Sigma_{\bY\bZ}$ : $\mathcal{H}_{\boldsymbol{\mathrm{z}}} \rightarrow \mathcal{H}_{\boldsymbol{\mathrm{y}}}$ satisfies the following equality for all $f \in \mathcal{H}_{\boldsymbol{\mathrm{z}}}$ 
and $g \in \mathcal{H}_{\boldsymbol{\mathrm{y}}}$:
\begin{align*}
\left\langle g, \Sigma_{\bY\bZ} f \right\rangle_{\mathcal{H}_{\boldsymbol{\mathrm{y}}}}
&= \mathbb{E}_{\bz\by} \left[ f(\bz)g(\by) \right] 
- \mathbb{E}_{\bz} \left[ f(\bz)\right]
  \mathbb{E}_{\by} \left[ g(\by)\right],
\end{align*}
where $\mathbb{E}_{\bz\by}$, $\mathbb{E}_{\bz}$, and $\mathbb{E}_{\by}$ 
denotes expectations over densities $p(\bz,\by)$, $p(\bz)$, and $p(\by)$, respectively.
Then, the \textit{conditional covariance operator} 
can be defined using cross-covariance operators as
\begin{align}
\Sigma_{\bY\bY|\bZ} = \Sigma_{\bY\bY} 
  - \Sigma_{\bY\bZ} \Sigma_{\bZ\bZ}^{-1} \Sigma_{\bZ\bY},	\label{KDR_cond_op}
\end{align}
where it is assumed that $\Sigma_{\bZ\bZ}^{-1}$ always exists.
The importance of the conditional covariance operator 
in supervised dimension reduction
lies in the following relations:
\begin{align}
\Sigma_{\bY\bY|\bZ} \geq \Sigma_{\bY\bY|\bX},	\label{KDR_prop1}
\end{align}
where the inequality refers to the partial order of self-adjoint operators,
and 
\begin{align}
\Sigma_{\bY\bY|\bZ} = \Sigma_{\bY\bY|\bX}
\Longleftrightarrow
(\bx \independent \by)~|~\bz.	\label{KDR_prop2}
\end{align}
These relations mean that 
the conditional independence can be achieved 
by finding a matrix $\bW$ which minimizes 
$\Sigma_{\bY\bY|\bZ}$ in the partial order of self-adjoint operators.
Based on this fact, KDR solves the following minimization problem:
\begin{align}
\min_{\bW \in \{ \bW | \bW\bW^\top = \boldsymbol{I}_{d_{\boldsymbol{\mathrm{z}}}} \}}
 \mathrm{Tr}\left[ 
\boldsymbol{G}_{{\boldsymbol{\mathrm{Y}}}}\left(\boldsymbol{G}_{{\boldsymbol{\mathrm{Z}}}} 
+ \lambda_n \boldsymbol{I}_n \right)^{-1} \right], \label{KDR_obj}
\end{align}
where $\lambda_n$ denotes a regularization parameter, 
$\boldsymbol{G}_{{\boldsymbol{\mathrm{Z}}}}$ and
$\boldsymbol{G}_{{\boldsymbol{\mathrm{Y}}}}$ denotes centered Gram matrices
with the kernels $k_{\boldsymbol{\mathrm{z}}}$ and $k_{\boldsymbol{\mathrm{y}}}$, respectively, 
and $\mathrm{Tr}\left[\cdot\right]$ denotes the trace of an operator.
A solution to this minimization problem 
is obtained by a gradient descent method.

KDR does not require any assumption on the data distribution 
and was shown to work well on various regression and classification tasks \citep{fukumizu2009}.
However, KDR has two disadvantages in practice.
The first disadvantage of KDR is that even though 
the kernel parameters and the regularization parameter
can heavily affect the performance, 
there seems to be no justifiable model selection method to choose these parameters so far.
Although it is always possible to choose these tuning parameters based on 
a criterion of a successive supervised learning method with cross-validation, 
this approach results in a nested loop of model selection 
for both KDR itself and the successive supervised learning method.
Moreover, this approach makes supervised dimension reduction 
depends on the successive supervised learning method
which is unfavorable in practice. 

The second disadvantage is that the optimization problem in Eq.\eqref{KDR_obj} 
is non-convex and may have many local solutions.
Thus, if the initial solution is not properly chosen, 
the performance of KDR may be unreliable.
A simple approach to cope with this problem is to choose 
the best solution with cross-validation
based on the successive supervised learning method,
but this approach makes supervised dimension reduction depends 
on the successive supervised learning method and is unfavorable.
A more sophisticated approach was considered in \cite{fukumizu2014}
which proposed to use a solution of a supervised dimension reduction method called 
\textit{gradient-based kernel dimension reduction} (gKDR)
as an initial solution for KDR.
However, it is not guarantee that 
gKDR always provide a good initial solution for KDR.

\subsection{Least-Squares Dimension Reduction}

The \textit{least-squares dimension reduction} 
(LSDR) \citep{DBLP:journals/neco/SuzukiS13}
is another supervised dimension reduction method
which does not require any assumption on the data distribution.
Similarly to KDR, LSDR aims to find a matrix 
$\bW$ which satisfies the conditional independence 
in Eq.\eqref{SDR_condition_ind}.
However, instead of the conditional covariance operators,
LSDR evaluates the conditional independence through
a statistical dependence measure.

LSDR utilizes a statistical dependence measure called 
\textit{squared-loss mutual information} (SMI).
SMI between random variables $\bz$ and $\by$ is defined as
\begin{align}
\mathrm{SMI}(Z,Y) = \frac{1}{2} \iint p(\bz)p(\by)
\left(\frac{p(\bz,\by)}{p(\bz)p(\by)} - 1\right)^2 
\dz\dy. \label{SMI_ZY}
\end{align}
$\mathrm{SMI}(Z,Y)$ is always non-negative and equals 
to zero if and only if $\bz$ and $\by$ are statistically independent, 
i.e., $p(\bz,\by) = p(\bz)p(\by)$.
The important properties of SMI in supervised dimension reduction
are the following relations:
\begin{align*}
\mathrm{SMI}(Z,Y) \leq \mathrm{SMI}(X,Y),
\end{align*}
and 
\begin{align*}
\mathrm{SMI}(Z,Y) = \mathrm{SMI}(X,Y)
\Longleftrightarrow
(\bx \independent \by)~|~\bz.
\end{align*}
Thus, the conditional independence can be achieved
by finding a matrix $\bW$ which maximizes $\mathrm{SMI}(Z,Y)$.
Since $\mathrm{SMI}(Z,Y)$ is typically unknown, 
it is estimated by the \textit{least-squares mutual information} 
\citep{DBLP:journals/bmcbi/SuzukiSKS09} method which directly estimates the density ratio 
$\frac{p(\bz,\by)}{p(\bz)p(\by)}$ 
without performing any density estimation. 
Then, LSDR solves the following maximization problem:
\begin{align}
\max_{\bW \in \{ \bW | \bW\bW^\top = \boldsymbol{I}_{d_{\boldsymbol{\mathrm{z}}}} \}}
 \widehat{\mathrm{SMI}}(Z,Y),	\label{SMI_opt}
\end{align}
where $\widehat{\mathrm{SMI}}(Z,Y)$ denotes the estimated SMI.
The solution to this maximization problem is obtained by a
gradient ascent method. 
Note that this maximization problem is non-convex and may have many local solutions.

LSDR does not require any assumption on the data distribution, 
similarly to dMAVE and KDR.
However, the significant advantage of LSDR over dMAVE and KDR
is that LSDR can perform model selection via cross-validation 
and avoid a poor local solution 
without requiring any successive supervised learning method.
This is a favorable property as a supervised dimension reduction method.

However, a disadvantage of LSDR is that
the density ratio function $\frac{p(\bz,\by)}{p(\bz)p(\by)}$
can be highly fluctuated, especially when the data contains outliers.
Since it is typically difficult to accurately estimate 
a highly fluctuated function, 
LSDR could be unreliable in the presence of outliers.

Next, we consider a supervised dimension reduction 
approach based on quadratic mutual information 
which can overcome the disadvantages of the existing methods.

\section{Quadratic Mutual Information}
\label{section:QMI}
In this section, we briefly introduce quadratic mutual information 
and discuss how it can be used to perform robust supervised dimension reduction.

\subsection{Quadratic Mutual Information and Mutual Information}
\textit{Quadratic mutual information} (QMI) 
is a measure for statistical dependency between random variables \citep{DBLP:journals/vlsisp/PrincipeXZF00},
and is defined as
\begin{align}
\mathrm{QMI}(Z,Y) 
= \frac{1}{2}\iint \left( p(\bz,\by) - p(\bz)p(\by)\right)^2 \dz\dy. \label{QMI_ZY}
\end{align}
$\mathrm{QMI}(Z,Y)$ is always non-negative
and equals to zero if and only if $\bz$ and $\by$ 
are statistically independent, i.e., $p(\bz,\by) = p(\bz)p(\by)$.
Such a property of QMI is similar to that
of the ordinary \textit{mutual information} (MI), which is defined as
\begin{align}
\mathrm{MI}(Z,Y) 
= \iint p(\bz,\by) \log\left( \frac{p(\bz,\by)}{p(\bz)p(\by)}\right) \dz\dy.
\label{MI_ZY}
\end{align}
The essential difference between QMI and MI is the discrepancy measure. 
$\mathrm{QMI}(Z,Y)$  is the $L_2$ distance between $p(\bz,\by)$ and $p(\bz)p(\by)$,
while $\mathrm{MI}(Z,Y)$ is the \textit{Kullback-Leibler} (KL) divergence \citep{Kullback51klDivergence}.

MI has been studied and applied to many data analysis tasks
\citep{Cover:1991:EIT:129837}.
Moreover, an efficient method to estimate MI from data 
is also available \citep{DBLP:journals/jmlr/SuzukiSSK08}.
However, MI is not always the optimal choice for measuring statistical dependence  
because it is not robust against outliers.
An intuitive explanation is that 
MI contains the log function and the density ratio:
the value of logarithm can be highly sharp near zero,
and density ratio can be highly fluctuated and diverge to infinity.
Thus, the value of MI tends to be unstable and unreliable 
in the presence of outliers.
In contrast, QMI does not contain the log function and the density ratio,
and thus QMI should be more robust against outliers than MI.

Another explanation of the robustness of QMI and MI 
can be understood based on their discrepancy measures.
Both $L_2$ distance (QMI) and KL divergence (MI) 
can be regarded as members of a more general divergence class 
called the \textit{density power} divergence \citep{Basu1998}:
\begin{align}
\mathrm{DP}_{\alpha}(p\|q) 
= \int\left(  p(\bx)^{1+\alpha} - 
\left(1 + \frac{1}{\alpha} \right) p(\bx)q(\bx)^{\alpha} 
+\frac{1}{\alpha}q(\bx)^{1+\alpha} \right) \dx, \label{DP_class}
\end{align}
where $\alpha > 0$.
Based on this divergence class, the $L_2$ distance and the 
KL divergence can be obtained by setting $\alpha = 1$ and $\alpha \rightarrow 0$, respectively.
As discussed in \cite{Basu1998}, the parameter $\alpha$ controls the robustness against outliers of the divergence, where a large value of $\alpha$ indicates high robustness.
This means that the $L_2$ distance ($\alpha = 1$)
is more robust against outliers than the KL divergence ($\alpha \rightarrow 0$).

In supervised dimension reduction, 
robustness against outliers is an important requirement
because outliers often make supervised dimension reduction methods to work poorly.
Thus, developing a supervised dimension reduction method 
based on QMI is an attractive approach 
since QMI is robust against outliers.
This QMI-based supervised dimension reduction method 
is performed by finding a matrix $\bW^*$
which maximizes $\mathrm{QMI}(Z, Y)$:
\begin{align*}
\bW^* = \argmax_{\bW \in \{ \bW | \bW\bW^\top 
= \boldsymbol{I}_{d_{\boldsymbol{\mathrm{z}}}} \}} 
\mathrm{QMI}(Z,Y).
\end{align*}
The motivation is that,
if $\mathrm{QMI}(Z, Y)$ is maximized
then $\bz$ and $\by$ are maximally dependent on each other,
and thus we may disregard $\bx$ 
with a minimal loss of information about $\by$.

Since $\mathrm{QMI}(Z, Y)$ is typically unknown,
it needs to be estimated from data.
Below, we review existing QMI estimation methods 
and then discuss a weakness of performing 
supervised dimension reduction using these QMI estimation methods.

\subsection{Existing QMI Estimation Methods}
We review two QMI estimation methods which estimate 
$\mathrm{QMI}(Z,Y)$ from the given data.
The first method estimates QMI through density estimation,
and the second method estimates QMI through density difference estimation.

\subsubsection{QMI Estimator based on Density Estimation}
Expanding Eq.\eqref{QMI_ZY} allows us to express
 $\mathrm{QMI}(Z,Y)$ as
\begin{align}
\mathrm{QMI}(Z,Y) 
&= \frac{1}{2}\iint \left( p(\bz,\by)^2 
- 2p(\bz,\by)p(\bz)p(\by) 
+ p(\bz)^2 p(\by)^2 \right) \dz\dy. \label{QMI_ZY_expand}
\end{align}
A naive approach to estimate $\mathrm{QMI}(Z,Y)$  is 
to separately estimate the unknown densities 
$p(\bz,\by)$, $p(\bz)$, and $p(\by)$ by density estimation methods such as \textit{kernel density estimation} (KDE) \citep{Silverman86},
and then plug the estimates into Eq.\eqref{QMI_ZY_expand}. 

Following this approach, the KDE-based QMI estimator 
has been studied and applied to many problems
such as \textit{feature extraction for classification} \citep{DBLP:journals/jmlr/Torkkola03, DBLP:journals/vlsisp/PrincipeXZF00}, 
\textit{blind source separation} \citep{DBLP:journals/vlsisp/PrincipeXZF00}, 
and \textit{image registration} \citep{Atif2003}.
Although this density estimation based approach was shown to work well, 
accurately estimating densities for 
high-dimensional data is known 
to be one of the most challenging tasks \citep{DBLP:books/vapnik1998}.
Moreover, the densities contained in Eq.\eqref{QMI_ZY_expand} 
are estimated independently 
without regarding the accuracy of the QMI estimator.
Thus, even if each density is accurately estimated,
the QMI estimator obtained from these density estimates 
does not necessarily give an accurate QMI. 
An approach to mitigate this problem is to consider density estimators 
which their combination minimizes the estimation error of QMI.
Although this approach shows better performance 
than the independent density estimation approach, 
it still performs poorly in high-dimensional problems
\citep{DBLP:journals/neco/SugiyamaKSPLT13}.

\subsubsection{Least-Squares QMI}
To avoid the separate density estimation, an alternative method called 
\textit{least-squares QMI} (LSQMI) \citep{DBLP:journals/ieicet/SainuiS13} was proposed. Below, we briefly review the LSQMI method.

First, notice that $\mathrm{QMI}(Z,Y)$ can be 
expressed in term of the density difference as
\begin{align}
\mathrm{QMI}(Z,Y) 
&= \frac{1}{2}\iint f(\bz,\by)^2 \dz\dy, \label{QMI_f}
\end{align}
where 
\begin{align*}
f(\bz, \by) = p(\bz,\by) - p(\bz)p(\by).
\end{align*}
The key idea of LSQMI is to directly estimate 
the density difference $f(\bz,\by)$ without going through any density estimation by the procedure of the \textit{least-squares density difference} \citep{DBLP:journals/neco/SugiyamaKSPLT13}.
Letting $d(\bz,\by)$ be a model of the density difference,
LSQMI learns $d(\bz,\by)$ so that it is fitted 
to the true density difference under the squared loss:
\begin{align*}
\frac{1}{2}\iint \left( d(\bz,\by) - f(\bz,\by) \right)^2 \dz\dy.
\end{align*}
By expanding the integrand, we obtain
\begin{align*}
\frac{1}{2}\iint d(\bz,\by)^2 \dz\dy
- \iint d(\bz,\by) f(\bz,\by) \dz\dy
+ \frac{1}{2} \iint f(\bz,\by)^2 \dz\dy.
\end{align*}
Since the last term is a constant w.r.t. the model $d(\bz,\by)$, 
we omit it and obtain the following criterion:
\begin{align}
\frac{1}{2}\iint d(\bz,\by)^2 \dz\dy
- \iint d(\bz,\by) f(\bz,\by) \dz\dy. \label{LSQMI_loss}
\end{align}
Then, the density difference estimator $\widehat{d}(\bz,\by)$ 
is obtained as the solution of the following minimization problem:
\begin{align}
\widehat{d} = \argmin_{d}
\left[ \frac{1}{2}\iint d(\bz,\by)^2 \dz\dy
- \iint d(\bz,\by) f(\bz,\by) \dz\dy \right]. \label{LSQMI_min}
\end{align}

The solution of the minimization problem in Eq.\eqref{LSQMI_min} 
depends on the choice of the model $d(\bz,\by)$.
LSQMI employs the following linear-in-parameter model
\begin{align*}
d(\bz,\by) = \balpha^\top \bpsi(\bz,\by),
\end{align*}
where $\balpha$ is a parameter vector and 
$\bpsi(\bz,\by)$ is a basis function vector.
For this model, finding the solution of Eq.\eqref{LSQMI_min} is equivalent to solving
\begin{align*}
\min_{\balpha}\left[
\frac{1}{2}\balpha^\top \boldsymbol{D} \balpha - \balpha^\top \boldsymbol{q} \right],
\end{align*}
where 
\begin{align}
\boldsymbol{D} 
&= \iint \bpsi(\bz,\by) \bpsi(\bz,\by)^\top \dz\dy, \\ 
\boldsymbol{q} 
&= \iint \bpsi(\bz,\by) f(\bz,\by) \dz\dy \notag \\
&= \iint \bpsi(\bz,\by) p(\bz,\by) \dz\dy -  
   \iint \bpsi(\bz,\by) p(\bz)p(\by) \dz\dy. \label{LSQMI_q_vec}
\end{align}
By approximating the expectation over the densities $p(\bz,\by)$, $p(\bz)$, 
and $p(\by)$ with sample averages, 
we obtain the following empirical minimization problem
\begin{align*}
\min_{\balpha}\left[
\frac{1}{2}\balpha^\top \boldsymbol{D} \balpha 
- \balpha^\top \widehat{\boldsymbol{q}} \right],
\end{align*}
where $\widehat{\boldsymbol{q}}$ is the sample approximation of Eq.\eqref{LSQMI_q_vec}:
\begin{align*}
\widehat{\boldsymbol{q}}
&= \frac{1}{n}\sum_{i=1}^n \bpsi(\bz_i,\by_i)
- \frac{1}{n^2}\sum_{i,j=1}^n \bpsi(\bz_i,\by_j).
\end{align*}
By including the $L_2$ regularization term, we obtain
\begin{align*}
\bhalpha = 
\argmin_{\balpha}\left[
\frac{1}{2}\balpha^\top \boldsymbol{D} \balpha 
- \balpha^\top \widehat{\boldsymbol{q}} 
+ \frac{\lambda}{2} \balpha^\top\balpha\right],
\end{align*}
where $\lambda \geq 0$ is the regularization parameter.
Then, the solution is obtained analytically as
\begin{align}
\bhalpha = \left( \boldsymbol{D} 
   + \lambda \boldsymbol{I}\right)^{-1} \widehat{\boldsymbol{q}}. \label{LSQMI_halpha}
\end{align}
Therefore, the density difference estimator is obtained as
\begin{align*}
\widehat{d}(\bz,\by) = \bhalpha^\top \bpsi(\bz,\by).
\end{align*}
Finally, QMI estimator is obtained by substituting the density difference estimator into Eq.\eqref{QMI_f}.
A direct substitution yields two possible QMI estimators:
\begin{align}
\widehat{\mathrm{QMI}}(Z,Y) 
&= \frac{1}{2} \bhalpha^\top \widehat{\boldsymbol{q}}, \label{LSQMI_1} \\
\widehat{\mathrm{QMI}}(Z,Y) 
&= \frac{1}{2} \bhalpha^\top \boldsymbol{D} \bhalpha. \label{LSQMI_2}
\end{align}
However, it was shown in \cite{DBLP:journals/neco/SugiyamaKSPLT13} that 
a linear combination of the two estimators defined as
\begin{align}
\widehat{\mathrm{QMI}}(Z,Y) 
&= \bhalpha^\top \widehat{\boldsymbol{q}} - \frac{1}{2} \bhalpha^\top \boldsymbol{D} \bhalpha, \label{LSQMI_3}
\end{align}
provides smaller bias and is a more appropriate QMI estimator.

As shown above, LSQMI avoids multiple-step density estimation 
by directly estimating the density difference contained in QMI.
It was shown that such direct estimation procedure tends to be more accurate than multiple-step estimation \citep{DBLP:journals/neco/SugiyamaKSPLT13}.
Moreover, LSQMI is able to objectively choose the tuning parameter contained
in the basis function $\bpsi(\bz,\by)$ and the regularization parameter $\lambda$ based on cross-validation.
This property allows LSQMI to solve challenging tasks such as
\textit{clustering} \citep{DBLP:journals/ieicet/SainuiS13} 
and \textit{unsupervised dimension reduction} \citep{DBLP:journals/ieicet/SainuiS14}
in an objective way.

\subsection{Supervised Dimension Reduction via LSQMI}
Given an efficient QMI estimation method such as LSQMI,  
supervised dimension reduction can be performed by finding a matrix $\bW^*$ defined as
\begin{align}
\bW^* = \argmax_{\bW \in \{ \bW | \bW\bW^\top = \boldsymbol{I}_{d_{\boldsymbol{\mathrm{z}}}} \}} 
\widehat{\mathrm{QMI}}(Z, Y). \label{SDR_QMI}
\end{align}
A straightforward approach to solving Eq.\eqref{SDR_QMI} 
is to perform the gradient ascent:
\begin{align*}
\bW \leftarrow \bW + t 
\frac{\partial \widehat{\mathrm{QMI}}(Z, Y)}{\partial \bW},
\end{align*}
where $t>0$ denotes the step size.
The update formula means that the essential point of 
the QMI-based supervised dimension reduction method is not the accuracy of the QMI estimator,
but the accuracy of the estimator of the derivative of the QMI. 
Thus, the existing LSQMI-based approach which first estimates QMI 
and then compute the derivatives of the QMI estimator 
is not necessarily appropriate 
since an accurate estimator of QMI does not 
necessarily mean that its derivative is an accurate estimator of the derivative of QMI.
Next, we describe our proposed method which overcomes this problem.

\section{Derivative of Quadratic Mutual Information}
\label{section:LSQMID}
To cope with the weakness of the QMI estimation methods 
when performing supervised dimension reduction,
we propose to \textit{directly} estimate the derivative of QMI without estimating QMI itself.

\subsection{Direct Estimation of the Derivative of Quadratic Mutual Information}

From Eq.\eqref{QMI_f}, the derivative of the ${\mathrm{QMI}}(Z,Y)$ 
w.r.t.~the $(\ell,\ell')$-th element of $\bW$ can be expressed by
\footnote{Throughout this section, we use ${\mathrm{QMI}}(\bW)$ 
instead of ${\mathrm{QMI}}(Z,Y)$ 
when we consider its derivative for notational convenience. 
However, they still represent the QMI between random variables $\bz$ and $\by$.}
\begin{align}
\frac{\partial \mathrm{QMI}(\bW)}{\partial W_{\ell,\ell'}}
 &= \frac{\partial}{\partial W_{\ell,\ell'}} 
 \left( \frac{1}{2}\iint f(\bz,\by)^2\dz\dy \right) \notag \\
 &= \iint f(\bz,\by) \frac{\partial f(\bz,\by)}{\partial W_{\ell,\ell'}}\dz\dy \notag \\
 &= \iint f(\bz,\by) \frac{\partial f(\bz,\by)}{\partial \bz}^\top 
    \frac{\partial \bz}{\partial W_{\ell,\ell'}}\dz\dy \notag \\
 &= \iint p(\bz,\by) \frac{\partial f(\bz,\by)}{\partial \bz}^\top 
    \frac{\partial \bz}{\partial W_{\ell,\ell'}}\dz\dy \notag \\
 &\phantom{=} - \iint p(\bz)p(\by) \frac{\partial f(\bz,\by)}{\partial \bz}^\top 
    \frac{\partial \bz}{\partial W_{\ell,\ell'}}\dz\dy, \label{QMID}
\end{align}
where in the second line we assume that the order of the derivative and the integration is interchangeable.
By approximating the expectations over the densities $p(\bz,\by)$,
$p(\bz)$, and $p(\by)$ with sample averages, 
we obtain an approximation of the derivative of QMI as
\begin{align}
\frac{\widehat{\partial \mathrm{QMI}}(\bW)}{\partial W_{\ell,\ell'}}
 &= \sum_{i=1}^n\frac{\partial f(\bz_i,\by_i)}{\partial \bz}^\top 
    \frac{\partial \bz_i}{\partial W_{\ell,\ell'}}
  - \sum_{i,j=1}^n\frac{\partial f(\bz_i,\by_j)}{\partial \bz}^\top 
    \frac{\partial \bz_i}{\partial W_{\ell,\ell'}}. \label{QMID_h}
\end{align} 
Note that since 
$z^{(\ell)} = \sum_{\ell'=1}^{d_{\boldsymbol{\mathrm{x}}}} W_{\ell, \ell'} x^{(\ell')}$, 
we have that $\frac{\partial \bz}{\partial W_{\ell,\ell'}}$ is the 
$d_{\boldsymbol{\mathrm{z}}}$-dimensional vector with zero everywhere 
except at the $\ell$-th dimension which has value $x^{(\ell')}$.
Hence, Eq.\eqref{QMID_h} can be simplified as 
\begin{align}
\frac{\widehat{\partial \mathrm{QMI}}(\bW)}{\partial W_{\ell,\ell'}} 
&= \sum_{i=1}^n\frac{\partial f(\bz_i,\by_i)}{\partial z^{(\ell)}}x_i^{(\ell')}
-\sum_{i,j=1}^n\frac{\partial f(\bz_i,\by_j)}{\partial z^{(\ell)}}x_i^{(\ell')}. \label{QMI_hat}
\end{align} 
This means that the derivative of $\mathrm{QMI}(Z,Y)$ w.r.t.~$\bW$
can be obtained once we know the derivatives 
of the density difference w.r.t.~$z^{(\ell)}$ for all 
$\ell \in \left\lbrace 1, \dots, d_{\boldsymbol{\mathrm{z}}} \right\rbrace$.
However, these derivatives are often unknown 
and need to be estimated from data.
Below, we first discuss existing approaches and their drawbacks.
Then we propose our approach which can overcome the drawbacks.

\subsection{Existing Approaches to Estimate the Derivative of the Density Difference}
Our current goal is to obtain the derivative of the density difference w.r.t.~$z^{(\ell)}$ which can be rewritten as
\begin{align}
\frac{\partial f(\bz,\by)}{\partial z^{(\ell)}}
&= \frac{\partial p(\bz,\by)}{\partial z^{(\ell)}} 
  - \frac{\partial p(\bz)}{\partial z^{(\ell)}} p(\by). \label{DDD_l}
\end{align}
All terms in Eq.\eqref{DDD_l} are unknown in practice and need to be estimated from data.
There are three existing approaches to estimate them.

\begin{description}
\item[(A) Density estimation] \hfill \\
Separately estimate the densities $p(\bz,\by)$, $p(\bz)$, and $p(\by)$ by, e.g., 
\textit{kernel density estimation}. 
Then estimate the right-hand side of Eq.\eqref{DDD_l} as
\begin{align*}
\frac{\partial \widehat{p}(\bz,\by)}{\partial z^{(\ell)}} 
  - \frac{ \partial \widehat{p}(\bz)}{\partial z^{(\ell)}} \widehat{p}(\by),
\end{align*}
where $\widehat{p}(\bz,\by)$, $\widehat{p}(\bz)$, and $\widehat{p}(\by)$ 
denote the estimated densities.

\item[(B) Density derivative estimation] \hfill \\
Estimate the density $p(\by)$ by e.g., kernel density estimation. 
Next, separately estimate the densities derivative 
$\frac{\partial p(\bz,\by)}{\partial z^{(\ell)}}$ 
and $\frac{\partial p(\bz)}{\partial z^{(\ell)}}$ by, e.g., 
the method of \textit{mean integrated square error for derivatives} 
\citep{DBLP:conf/aistats/SasakiNS15},
which can estimate the density derivative 
without estimating the density itself.
Then estimate the right-hand side of Eq.\eqref{DDD_l} as
\begin{align*}
\frac{\widehat{\partial p}(\bz,\by)}{\partial z^{(\ell)}} 
  - \frac{ \widehat{\partial p}(\bz)}{\partial z^{(\ell)}} \widehat{p}(\by),
\end{align*}
where $\widehat{p}(\by)$ denotes the estimated density, and
$\frac{\widehat{\partial p}(\bz,\by)}{\partial z^{(\ell)}}$ and 
$\frac{\widehat{\partial p}(\bz)}{\partial z^{(\ell)}}$ 
denote the (directly) estimated density derivatives.

\item[(C) Density difference estimation] \hfill \\
Estimate the density difference $f(\bz,\by)$ by e.g., \textit{least-squares density difference} \citep{DBLP:journals/neco/SugiyamaKSPLT13},
which can estimate the density difference 
without estimating the densities themselves.
Then estimate the left-hand side of Eq.\eqref{DDD_l} as
\begin{align*}
\frac{\partial \widehat{f}(\bz,\by)}{\partial z^{(\ell)}} ,
\end{align*}
where $\widehat{f}(\bz,\by)$ denotes the (directly) estimated density difference.
\end{description}

The problem of approaches (A) and (B) is that
they involve multiple estimation steps 
where some quantities are estimated first 
and then they are plugged into Eq.\eqref{DDD_l}.
Such multiple-step methods are not appropriate 
since each estimated quantity is obtained without 
regarding the others and the succeeding plug-in step using these estimates 
can magnify the estimation error contained in each estimated quantity.

On the other hand, approach (C) seems more promising than the previous two approaches since there is only one estimated quantity $f(\bz,\by)$.
However, it is still not the optimal approach due to the fact that an accurate estimator of the density difference does not necessarily means that its derivative is an accurate estimator of the derivative of the density difference.

To avoid the above problems, 
we propose a new approach which directly estimates the derivative of the density difference.

\subsection{Direct Estimation of the Derivative of the Density Difference}
We propose to estimate the derivative of the density difference 
w.r.t.~$z^{(\ell)}$ using a model $g_{\ell}(\bz,\by)$:
\begin{align*}
\frac{\partial f(\bz,\by)}{\partial z^{(\ell)}} \approx g_{\ell}(\bz,\by).
\end{align*}
The model $g_{\ell}(\bz,\by)$ is learned so that 
it is fitted to its corresponding derivative under the square loss:
\begin{align}
\frac{1}{2}\iint \left( g_{\ell}(\bz,\by) 
- \frac{\partial f(\bz,\by)}{\partial z^{(\ell)}} \right)^2\dz\dy.
\end{align}
By expanding the square, we obtain
\begin{align*}
\frac{1}{2}\iint  g_{\ell}(\bz,\by)^2 \dz\dy
- \iint g_{\ell}(\bz,\by) \frac{\partial f(\bz,\by)}{\partial z^{(\ell)}} \dz\dy 
+ \frac{1}{2} \iint 
\left(\frac{\partial f(\bz,\by)}{\partial z^{(\ell)}} \right)^2 \dz\dy.
\end{align*}
Since the last term is a constant w.r.t.~the model $g_{\ell}(\bz,\by)$, 
we omit it and obtain the following criterion:
\begin{align}
 \frac{1}{2}\iint g_{\ell}(\bz,\by) ^2 \dz\dy
   - \iint g_{\ell}(\bz,\by) \frac{\partial f(\bz,\by)}{\partial z^{(\ell)}} \dz\dy. \label{J_l}
\end{align}
The second term is intractable due to the 
unknown derivative of the density difference.
To make this term tractable, 
we use \textit{integration by parts} \citep{integrationbyparts1983} to obtain the following:
\begin{align}
&\iint \left[ g_{\ell}(\bz,\by) f(\bz,\by)\right]_{z^{(\ell)} 
= -\infty}^{z^{(\ell)} = \infty} \dz_{\backslash z^{(\ell)}} \dy \notag \\
&\quad\quad = \iint f(\bz,\by) \frac{ \partial g_{\ell}(\bz,\by)}{\partial z^{(\ell)}} \dz\dy 
 + \iint g_{\ell}(\bz,\by) \frac{ \partial f(\bz,\by)}{\partial z^{(\ell)}} \dz\dy, \label{int_by_parts}
\end{align}
where $\int \cdot \dz_{\backslash z^{(\ell)}}$ 
denotes an integration over $\bz$ except for the $\ell$-th element.
Here, we require 
\begin{align}
\left[ g_{\ell}(\bz,\by) f(\bz,\by)\right]
_{z^{(\ell)} = -\infty}^{z^{(\ell)} = \infty} = 0, \label{assumption}
\end{align}
which is a mild assumption since 
the tails of the density difference $p(\bz,\by) - p(\bz)p(\by)$ 
often vanish when $z^{(\ell)}$ approaches infinity.
Applying the assumption to the left-hand side of Eq.\eqref{int_by_parts}
allows us to express Eq.\eqref{J_l} as
\begin{align*}
\frac{1}{2}\iint g_{\ell}(\bz,\by)^2 \dz\dy 
+ \iint f(\bz,\by) \frac{ \partial g_{\ell}(\bz,\by)}{\partial z^{(\ell)}}\dz\dy.
\end{align*}
Then, the estimator $\widehat{g}_{\ell}(\bz,\by)$ is obtained 
as a solution of the following minimization problem:
\begin{align}
\widehat{g}_{\ell} 
&= \argmin_{g_{\ell}}\left[ \frac{1}{2}\iint g_{\ell}(\bz,\by)^2 \dz\dy
   + \iint  f(\bz,\by) \frac{\partial g_{\ell}(\bz,\by)}{\partial z^{(\ell)}}\dz\dy \right]. \label{g_min}
\end{align}

The solution of Eq.\eqref{g_min} depends on the choice of the model.
Let us employ the following linear-in-parameter model as $g_{\ell}(\bz,\by)$:
\begin{align}
g_{\ell}(\bz,\by) = \btheta_{\ell}^\top \bvphi_{\ell}(\bz,\by), \label{g_model}
\end{align}
where $\btheta_{\ell}$ is a parameter vector 
and $\bvphi_{\ell}(\bz,\by)$ is a basis function vector 
whose practical choice will be discussed later in detail.
For this model, finding the solution of Eq.\eqref{g_min} 
is equivalent to solving
\begin{align}
\min_{\btheta_{\ell}}\left[ 
\frac{1}{2} \btheta_{\ell}^\top \bH_{\ell} \btheta_{\ell} 
+ \btheta_{\ell}^\top \bh_{\ell} \right], 
\label{J_l_theta}
\end{align}
where we define
\begin{align}
\bH_{\ell} 
&= \iint \bvphi_{\ell}(\bz,\by)\bvphi_{\ell}(\bz,\by)^\top \dz\dy, \label{H_l} \\
\bh_{\ell} &=  \iint f(\bz,\by )\frac{\partial \bvphi_{\ell}(\bz,\by)}{\partial z^{(\ell)}} \dz\dy \notag \\
&= \iint p(\bz,\by )\frac{\partial \bvphi_{\ell}(\bz,\by)}{\partial z^{(\ell)}} \dz\dy 
- \iint p(\bz)p(\by)\frac{\partial \bvphi_{\ell}(\bz,\by)}{\partial z^{(\ell)}} \dz\dy. \label{h_l}
\end{align}
By approximating the expectation over the densities $p(\bz,\by)$, $p(\bz)$, and $p(\by)$ with sample averages, 
we obtain the following empirical minimization problem:
\begin{align}
\min_{\btheta_{\ell}}\left[ 
\frac{1}{2} \btheta_{\ell}^\top \bH_{\ell} \btheta_{\ell} 
+ \btheta_{\ell}^\top \bhh_{\ell} \right], 
\end{align}
where $\bhh_{\ell}$ is the sample approximation of Eq.\eqref{h_l}:
\begin{align}
\bhh_{\ell}
&= \frac{1}{n} \sum_{i=1}^n \frac{\partial \bvphi_{\ell}(\bz_i,\by_i)}{\partial z^{(\ell)}}
 - \frac{1}{n^2} \sum_{i,j=1}^n \frac{\partial \bvphi_{\ell}(\bz_i,\by_j)}{\partial z^{(\ell)}}. \label{hh_l}
\end{align}
By including the $L_2$ regularization term 
to control the model complexity, we obtain
\begin{align}
\bhtheta_{\ell} 
&= \argmin_{\btheta_{\ell}}\left[ \frac{1}{2} \btheta_{\ell}^\top \bH_{\ell} \btheta_{\ell} 
   + \btheta_{\ell}^\top \bhh_{\ell} + \frac{\lambda_{\ell}}{2} 
   \btheta_{\ell}^\top \btheta_{\ell} \right], 
\end{align}
where $\lambda_{\ell} \ge 0$ denotes the regularization parameter.
This minimization problem is convex w.r.t.~the parameter $\btheta_{\ell}$,
and the solution can be obtained analytically as
\begin{align}
\bhtheta_{\ell} 
&= -\left( \bH_{\ell} + \lambda_{\ell} \boldsymbol{I} \right)^{-1} \bhh_{\ell}, \label{theta_hat}
\end{align}
where $\boldsymbol{I}$ denotes the identity matrix.
Finally, the estimator of the derivative of the density difference 
is obtained by substituting the solution into the model Eq.\eqref{g_model} as
\begin{align}
\widehat{g}_{\ell}(\bz,\by) = \bhtheta_{\ell}^\top \bvphi_{\ell}(\bz,\by). \label{gh_l}
\end{align}

Using this solution, an estimator of the derivative of QMI can be 
directly obtained by substituting Eq.\eqref{gh_l} into Eq.\eqref{QMI_hat} as
\begin{align}
\frac{\widehat{\partial \mathrm{QMI}}(\bW)}{\partial W_{\ell,\ell'}}
&= \frac{1}{n}\sum_{i=1}^n \bhtheta_{\ell}^\top \bvphi_{\ell}(\bz_i,\by_i) x^{(\ell')}_i
  - \frac{1}{n^2}\sum_{i,j=1}^n \bhtheta_{\ell}^\top \bvphi_{\ell}(\bz_i,\by_j) x^{(\ell')}_i.    \label{LSQMID}
\end{align}
We call this method the \textit{least-squares QMI derivative} (LSQMID).

\subsection{Basis Function Design}
As basis function $\bvphi_{\ell}(\bz,\by)$, we propose to use
\begin{align*}
\bvphi_{\ell}(\bz,\by) = \left[\varphi^{(1)}_{\ell}(\bz,\by), 
	\cdots, \varphi^{(b)}_{\ell}(\bz,\by) \right]^\top,
\end{align*}
where $b \leq n$.
First, let us define the $k$-th Gaussian function as 
\begin{align}
\phi_{\ell}^{(k)}(\bz, \by) = 
\exp\left( -\frac{ \| \bz - \bu_k \|^2 + \| \by - \bv_k\|^2}{2\sigma_{\ell}^2} \right), \label{phi_lk}
\end{align}
where $\bu_k$ and $\bv_k$ denote Gaussian centers chosen randomly from the data samples $\{\bz_i, \by_i\}_{i=1}^n$, 
and $\sigma_{\ell}$ denotes the Gaussian width.
We may use different Gaussian widths for $\bz$ and $\by$,
but this approach significantly increases the computation time 
for model selection which will be discussed in Section~\ref{cross_validation}.
In our implementation, 
we standardize each dimension of $\bx$ and $\by$ to have unit variance and zero mean,
and then use the common Gaussian width for both $\bz$ and $\by$.
We also set $b = \min(n, 200)$ in the experiments.

Based on the above Gaussian function, we propose to use the following function as the $k$-th basis
for the $\ell$-th model of the derivative of the density difference:
\begin{align}
\varphi_{\ell}^{(k)}(\bz,\by) 
&= \frac{\partial \phi_{\ell}^{(k)}(\bz,\by)}{\partial z^{(\ell)}}\notag \\
&= -\frac{1}{\sigma_{\ell}^2}(z^{(\ell)} - u_k^{(\ell)})\phi_{\ell}^{(k)}(\bz,\by). \label{vphi_lk}
\end{align}
This function is the derivative of the $k$-th Gaussian basis function 
w.r.t.~$z^{(\ell)}$.
A benefit of this basis function design is that 
the integral appeared in $\bH_{\ell}$ can be computed analytically.
Through some simple calculation, we obtain the $(k,k')$-th element of $\bH_{\ell}$ as follows:
\begin{align*}
H_{{\ell}}^{(k,k')} &= \frac{1}{\sigma_{\ell}^4} 
(\sqrt{\pi}\sigma_{\ell})^{d_{\boldsymbol{\mathrm{z}}} + d_\mathrm{\by}}
\exp\left(-\frac{ \|\bu_k - \bu_{k'}\|^2 - \| \bv_k - \bv_{k'}\|^2 }{4\sigma_{\ell}^2}\right)\\
&\quad \times \left(
u^{(\ell)}_k u^{(\ell)}_{k'} 
- \frac{(u^{(\ell)}_k +u^{(\ell)}_{k'})^2}{2} 
+ (\frac{u^{(\ell)}_k+u^{(\ell)}_{k'} }{2} )^2
+ \frac{\sigma_{\ell}^2}{2}
\right).
\end{align*}

As discussed in Section~\ref{section:SDR_LSQMID}, this basis function choice 
has further benefits when we develop a supervised dimension reduction method.

\subsection{Model Selection by Cross-Validation}
\label{cross_validation}
The practical performance of the LSQMID method depends on the choice of 
the Gaussian width $\sigma_{\ell}$ and 
the regularization parameter $\lambda_{\ell}$ 
included in the estimator $\widehat{g}_{\ell}(\bz,\by)$.
These tuning parameters can be objectively chosen by the
$K$-fold cross-validation (CV) procedure which is described below.

\begin{enumerate}

\item Divide the training data $\mathcal{D} = \{(\bx_i, \by_i)\}_{i=1}^n$
into $K$ disjoint subsets $\{\mathcal{D}_j\}_{j=1}^K$
with approximately the same size.
In the experiments, we choose $K = 5$.

\item For each candidate  
$M = (\tilde{\sigma}_{\ell}, \tilde{\lambda}_{\ell})$
and each subset $\mathcal{D}_j$, compute a solution
$\bhtheta_{\ell,M,\backslash j}$
by Eq.\eqref{theta_hat}
with the candidate $M$ and samples from
$\mathcal{D} \backslash \mathcal{D}_j$ 
(i.e., all data samples except samples in $\mathcal{D}_j$).

\item Compute the CV score of each candidate pair $M$ by
\begin{align*}
\mathrm{CV}_{\ell}(M) &= \frac{1}{K}\sum_{j=1}^K
\left[
\frac{1}{2} \bhtheta_{\ell,M, \backslash j}^\top \bH_{\ell,M} \bhtheta_{\ell,M, \backslash j}
 + \bhtheta_{\ell,M, \backslash j}^\top \bhh_{\ell,M,j}
 \right],
\end{align*}
where $\bhh_{\ell,M,j}$ denotes $\bhh_{\ell}$ 
computed from the candidate $M$ 
and samples in $\mathcal{D}_j$.

\item 
Choose the tuning parameter pair such that it minimizes the CV score as
\begin{align*}
(\widehat{\sigma}_{\ell}, \widehat{\lambda}_{\ell}) = \argmin_{M} \mathrm{CV}_{\ell}(M).
\end{align*}
\end{enumerate}


\section{Supervised Dimension Reduction via LSQMID}
\label{section:SDR_LSQMID}
In this section, we propose a supervised dimension reduction method 
based on the proposed LSQMID estimator.
\subsection{Gradient ascent via LSQMID}
Recall that our goal in supervised dimension reduction is to find the matrix $\bW^*$:
\begin{align}
\bW^* = \argmax_{\bW \in \{ \bW | \bW\bW^\top = \boldsymbol{I}_{d_{\boldsymbol{\mathrm{z}}}} \}} \mathrm{QMI}(Z,Y). \label{max_QMI}
\end{align}
A straightforward approach to find a solution of 
Eq.\eqref{max_QMI} using the proposed method is 
to perform gradient ascent as
\begin{align}
\bW \leftarrow \bW + t \frac{\widehat{\partial \mathrm{QMI}}(\bW)}{\partial \bW}, \label{gradient_ascent_eu}
\end{align}
where $t > 0$ denotes the step size.
It is known that choosing a good step size is a difficult task in practice \citep{Nocedal06}.
\textit{Line search} is an algorithm to choose a good step size
by finding a step size which satisfies 
certain conditions such as the \textit{Armijo rule} \citep{armijo1966}.
However, these conditions often require access to 
the objective value $\mathrm{QMI}(\bW)$ 
which is unavailable in our current setup
since the QMI derivative is directly estimated without estimating QMI.
Thus, if we want to perform line search, 
QMI needs to be estimated separately.
However, this is problematic 
since the estimation of the derivative of the QMI 
and the estimation of the QMI are performed independently 
without regard to the other, and thus they may not be consistent.
For example, the gradient 
$\frac{\widehat{\partial \mathrm{QMI}}(\bW)}{\partial \bW}$
,which is supposed to be an ascent direction,
may be regarded as a descent direction 
on the surface of the estimated QMI.
For such a case, the step size chosen by 
any line search algorithm is unreliable 
and the resulting $\bW$ may not be a good solution.

Below, we consider two approaches which can cope with this problem.

\subsection{QMI Approximation via LSQMID}
\label{section_dqmi_gr}
To avoid separate QMI estimation, we consider an approximated QMI 
which is obtained as a by-product of the proposed method.
Recall that the proposed method models 
the derivative of the density difference as
\begin{align*}
\frac{\partial f(\bz,\by)}{\partial z^{(\ell)}}
&\approx g_{\ell}(\bz,\by) \\
&= \btheta_{\ell}^\top \bvphi_{\ell}(\bz,\by) \\
&=\btheta_{\ell}^\top \frac{\partial \bphi_{\ell}(\bz,\by)}{\partial z^{(\ell)}}\\
&=\frac{\partial \left( \btheta_{\ell}^\top \bphi_{\ell}(\bz,\by) \right)}{\partial z^{(\ell)}}.
\end{align*}
This means that the density difference can be approximated by  
\begin{align}
\widetilde{f}_{\ell}(\bz,\by) = \bhtheta_{\ell}^\top \bphi_{\ell}(\bz,\by) + c_{\ell}, \label{f_approx_LSQMID}
\end{align}
where $c_{\ell}$ is an unknown quantity 
which is a constant w.r.t.~$z^{(\ell)}$.

In a special case where $d_{\boldsymbol{\mathrm{z}}} = 1$, 
we can use Eq.\eqref{f_approx_LSQMID}
to obtain a proper approximator of 
$\mathrm{QMI}(Z,Y)$ in a similar fashion to the LSQMI method.
To verify this, let us substitute Eq.\eqref{f_approx_LSQMID} 
into one of the $f(z,\by)$ in Eq.\eqref{QMI_f} to obtain
\begin{align*}
\widetilde{\mathrm{QMI}}(Z,Y) 
&=  \frac{1}{2}\iint f(z,\by) \widetilde{f}(z,\by) 
 \mathrm{d}z\dy \\
&= \frac{1}{2}\iint f(z,\by) 
\left(  \bhtheta^\top \bphi(z,\by) + c \right) \mathrm{d}z\dy \\
&= \frac{1}{2}\iint f(z,\by) 
 \bhtheta^\top \bphi(z,\by) \mathrm{d}z\dy 
+ \frac{1}{2}\iint f(z,\by) c \mathrm{d}z\dy \\
&= \frac{1}{2}\iint f(z,\by) 
 \bhtheta^\top \bphi(z,\by) \mathrm{d}z\dy,
\end{align*}
where the last line follows from
\begin{align*}
\iint f(z,\by) c \mathrm{d}z\dy
&= \iint p(z,\by) c \mathrm{d}z\dy 
- \iint p(z)p(\by) c \mathrm{d}z\dy \\
&= 0.
\end{align*}
By approximating the expectation with sample averages, 
we obtain a QMI approximator as
\begin{align}
\widetilde{\mathrm{QMI}}(Z,Y) 
&= \frac{1}{2n} \sum_{i=1}^n \bhtheta^\top \bphi(z_i,\by_i)
-\frac{1}{2n^2}\sum_{i,j=1}^n \bhtheta^\top \bphi(z_i,\by_j). \label{QMI_LSQMID}
\end{align}

The main advantage of using $\widetilde{\mathrm{QMI}}(Z,Y)$ 
is that it is obtained from the derivative estimation,
and thus should be consistent with the estimated derivative.
This allows us to perform line search for the 
gradient ascent in a consistent manner.
We may further improve the optimization procedure by considering 
an optimization problem over the \textit{Grassmann manifold}:
\begin{align}
\bW^* = \argmax_{\bW \in \mathrm{Gr}^{d_{\boldsymbol{\mathrm{x}}}}_{d_{\boldsymbol{\mathrm{z}}}}} 
\widetilde{\mathrm{QMI}}(Z,Y), \label{max_QMI_gr}
\end{align}
where $\mathrm{Gr}^{d_{\boldsymbol{\mathrm{x}}}}_{d_{\boldsymbol{\mathrm{z}}}}$ is defined as 
\begin{align*}
\mathrm{Gr}^{d_{\boldsymbol{\mathrm{x}}}}_{d_{\boldsymbol{\mathrm{z}}}} 
:= \{\bW \in \mathbb{R}^{d_{\boldsymbol{\mathrm{z}}} \times d_{\boldsymbol{\mathrm{x}}}} \ | \ \bW\bW^T = \boldsymbol{I}_{d_{\boldsymbol{\mathrm{z}}}}\}/\sim.
\end{align*}
That is, $\mathrm{Gr}^{d_{\boldsymbol{\mathrm{x}}}}_{d_{\boldsymbol{\mathrm{z}}}}$ 
is a set of $d_{\boldsymbol{\mathrm{z}}}$-by-$d_{\boldsymbol{\mathrm{x}}}$ 
orthonormal matrices whose rows span the same subspace.
This manifold optimization is more efficient than the original optimization  
since every step of the optimization always satisfies the orthonormal constraint,
and we no longer need to perform orthonormalization.
More details of manifold optimization can be found in \cite{AbsMahSep2008}.

Although the QMI approximation in Eq.\eqref{QMI_LSQMID} 
allows us to choose step size by line search in a consistent manner,
such an approximation is unavailable when $d_{\boldsymbol{\mathrm{z}}} > 1$.
Next, we consider an alternative optimization strategy 
which does not require an access to the QMI value.

\subsection{Fixed-Point Iteration}
\label{section_dqmi_fp}

To avoid the problem of choosing the step size 
which requires an access to the QMI value, 
we propose to use a fixed-point iteration for 
finding a solution of Eq.\eqref{max_QMI}.
Note that from the first order optimality condition,
a solution $\bW^*$ is a stationary point which satisfies 
\begin{align*}
\frac{\partial \mathrm{QMI}(\bW^*)}{\partial \bW} = \boldsymbol{0}_{d_{\boldsymbol{\mathrm{z}}},d_{\boldsymbol{\mathrm{x}}}},
\end{align*}
where $\boldsymbol{0}_{d_{\boldsymbol{\mathrm{z}}},d_{\boldsymbol{\mathrm{x}}}}$ 
denotes $d_{\boldsymbol{\mathrm{z}}}$-by-$d_{\boldsymbol{\mathrm{x}}}$ zero matrix.
By using the proposed basis function in Eq.\eqref{vphi_lk}, 
Eq.\eqref{LSQMID} can be expressed as
\begin{align}
\frac{\widehat{\partial \mathrm{QMI}}(\bW)}{\partial W_{\ell,\ell'}}
= F_1^{(\ell,\ell')} - F_2^{(\ell,\ell')} - W_{\ell,\ell'}F_3^{(\ell,\ell')}, \label{LSQMID_FP}
\end{align}
where we define
\begin{align*}
F_{1}^{(\ell,\ell')} 
&= \bhtheta_{\ell}^\top \left( \bu^{(\ell)} \odot
\left( \frac{1}{n}\sum_{i=1}^n \bphi_{\ell}(\bz_i,\by_i) x_i^{(\ell')}  
- \frac{1}{n^2}\sum_{i,j=1}^n \bphi_{\ell}(\bz_i,\by_j) x_i^{(\ell')}    
\right)\right)\sigma_{\ell}^{-2}, \\
F_2^{(\ell,\ell')} &= \sum_{m \neq {\ell'}}^{d_{\boldsymbol{\mathrm{x}}}} W_{\ell, m}
\bhtheta_{\ell}^\top 
\left(  \frac{1}{n}\sum_{i=1}^n \bphi_{\ell}(\bz_i,\by_i) x_i^{(m)} x_i^{(\ell')}  
-  \frac{1}{n^2}\sum_{i,j=1}^n \bphi_{\ell}(\bz_i,\by_j) x_i^{(m)}  x_i^{(\ell')}    
\right) \sigma_{\ell}^{-2}, \\
F_3^{(\ell,\ell')} &= \bhtheta_{\ell}^\top 
\left(  \frac{1}{n}\sum_{i=1}^n \bphi_{\ell}(\bz_i,\by_i) x_i^{(l')} x_i^{(\ell')}  
-  \frac{1}{n^2}\sum_{i,j=1}^n \bphi_{\ell}(\bz_i,\by_j) x_i^{(\ell')}  x_i^{(\ell')} \right) \sigma_{\ell}^{-2},
\end{align*}
with $\bu^{(\ell)}$ be the column vector of length $b$
consisting of the $\ell$-th dimension over all $\bu_k$
and the symbol $\odot$ represents the element-wise vector product.
Then, an approximated solution may be obtained by finding 
$W_{\ell,\ell'}$ for all $(\ell,\ell')$ such that the left-hand side of Eq.\eqref{LSQMID_FP} is zero.
This optimization strategy results in a 
fixed-point iteration for each dimension of $\bW$:
\begin{align*}
W_{\ell,\ell'} \leftarrow \frac{F_{1}^{(\ell,\ell')} - F_{2}^{(\ell,\ell')}}{F_{3}^{(\ell,\ell')} }. 
\end{align*}
Finally, we orthonormalize the solution after each iteration as
\begin{align*}
\bW \leftarrow \left( \bW\bW^\top \right)^{-\frac{1}{2}}\bW.
\end{align*}
In practice, we perform this orthonormalization 
only every several iterations for computational efficiency.

Note that the optimization problem in Eq.\eqref{max_QMI}
is non-convex and may have many local solutions.
To avoid obtaining a poor local optimal solution,
we perform the optimization
starting from several initial guesses 
and choose the solution which gives 
the maximum estimated QMI as the final solution.


\section{Experiments}
\label{section:experiments}
In this section, we demonstrate the usefulness of the proposed method
through experiments.
\subsection{Illustrative Experiment}
Firstly, we perform the following experiment to illustrate 
the usefulness of the proposed method in term of the QMI derivative estimation.
Let $\mathcal{N}(\boldsymbol{\mu}, \boldsymbol{\Sigma})$ 
denotes the Gaussian distribution with mean 
$\boldsymbol{\mu}$ and covariance $\boldsymbol{\Sigma}$.
Then, for $\epsilon \sim \mathcal{N}(0, 0.15^2)$,
we generate a dataset $\{(\bx_i,y_i)\}_{i=1}^n$ 
and a matrix $\bW$ as follows:
\begin{align*}
\bx &\sim \mathcal{N}(\boldsymbol{0}_2, \boldsymbol{I}_2), \\
y &= (x^{(1)})^2 + \epsilon, \\
\bW &= \begin{bmatrix}
\cos\theta & \sin\theta
\end{bmatrix},
\end{align*}
where $\boldsymbol{0}_2$ denotes a zero vector of length 2.
Thus we have $z  = x^{(1)}\cos\theta + x^{(2)}\sin\theta$.
The goal is to estimate 
\begin{align*}
\frac{\partial \mathrm{QMI}(Z,Y)}{\partial \theta} = 
\frac{\partial \mathrm{QMI}(Z,Y)}{\partial \bW}
\frac{\partial \bW}{\partial \theta}
\end{align*}
at different value of $\theta$.
Note that $\mathrm{QMI}(Z,Y)$ 
is maximized at $\theta = 0$, i.e., 
$\bW = \begin{bmatrix}1 & 0\end{bmatrix}$.

\begin{figure}[t]
  \begin{subfigure}[b]{0.50\linewidth}
  \centering
    \includegraphics[width=1\textwidth]{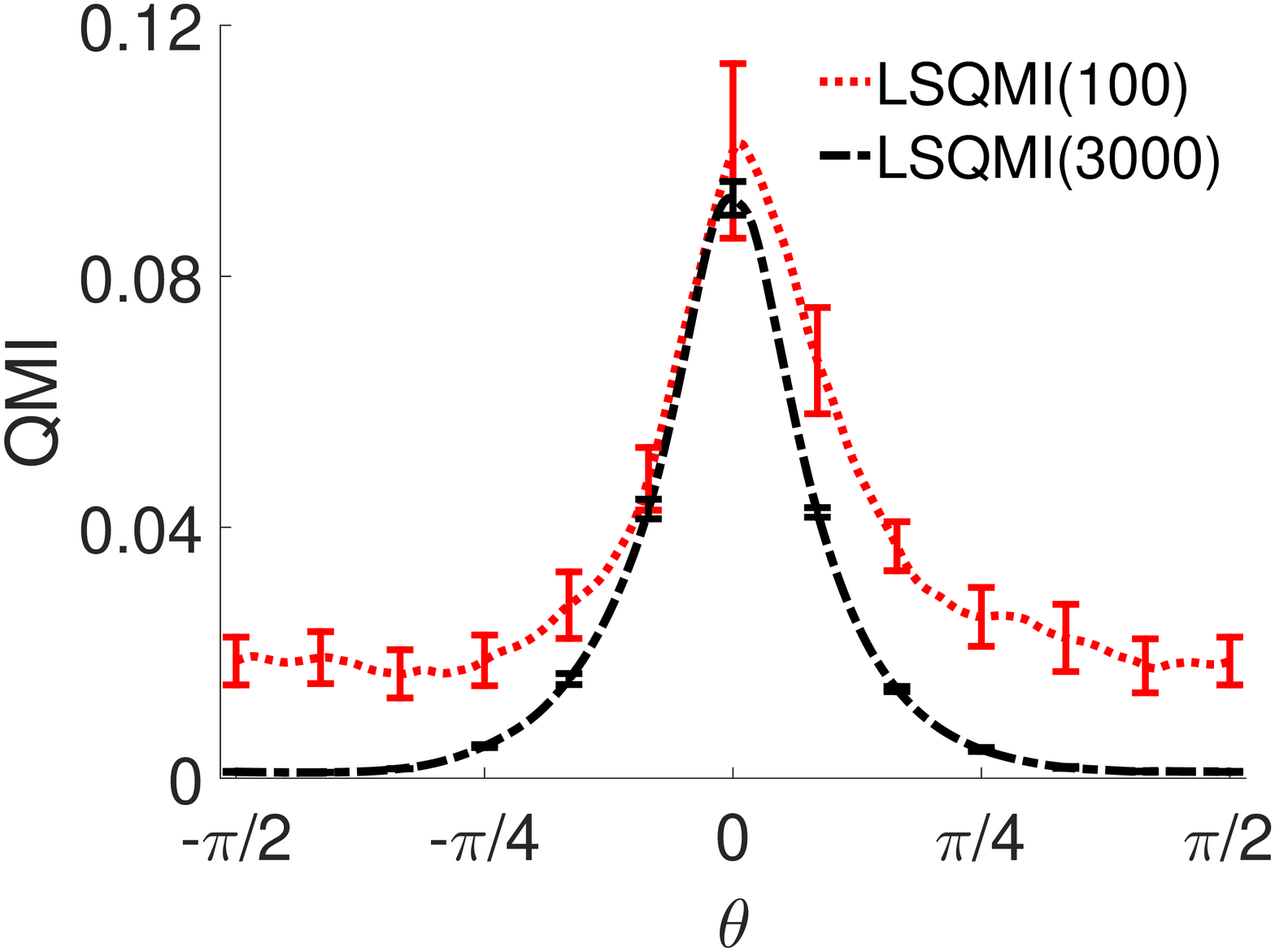}
    \subcaption{The averaged estimated QMI.}
    \label{fig_qmi_illust_avg}
  \end{subfigure}
  \begin{subfigure}[b]{0.50\linewidth}
  \centering
    \includegraphics[width=1\textwidth]{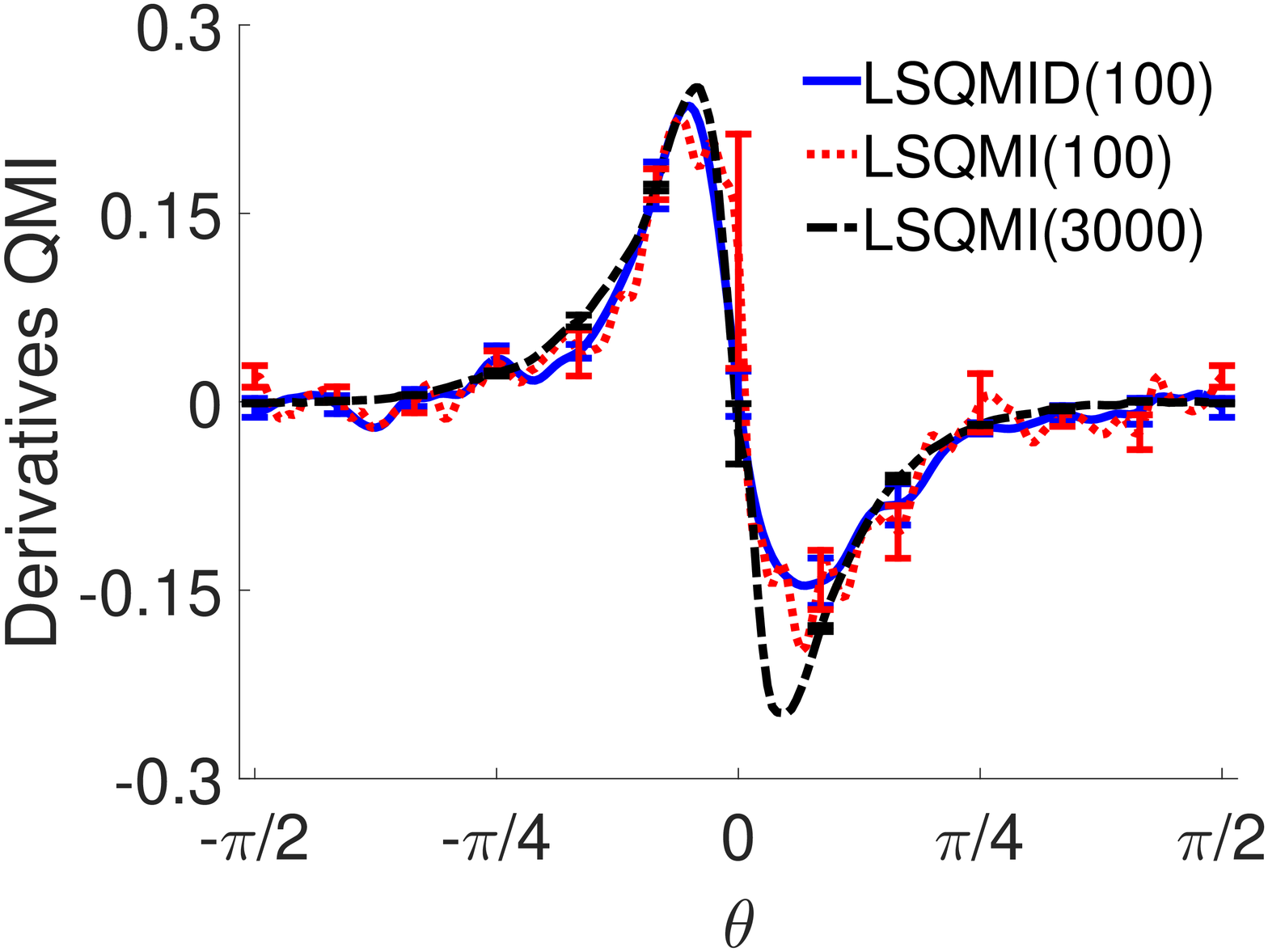}
    \subcaption{The averaged estimated derivative of QMI.}
    \label{fig_dqmi_illust_avg}
  \end{subfigure}
  \caption{The mean and standard error of the estimated $\mathrm{QMI}(Z,Y)$ and the estimated derivative of $\mathrm{QMI}(Z,Y)$ w.r.t.~$\theta$ over 20 experiment trials.}
\end{figure}

Figure~\ref{fig_qmi_illust_avg} shows 
the averaged value over 20 experiment trials 
of the estimated $\mathrm{QMI}(Z,Y)$ by LSQMI.
The vertical axis indicates the value of the estimated QMI 
and the horizontal axis indicates value of 
$\theta \in \left[ -\frac{\pi}{2}, \frac{\pi}{2} \right]$.
We use $n=3000$ and $n=100$ for estimating QMI
and denote the results by LSQMI(3000) and LSQMI(100), respectively.
We perform cross validation at $\theta = 0$ 
and use the chosen tuning parameters for all values of $\theta$.
The result shows that LSQMI accurately 
estimates $\mathrm{QMI}(Z,Y)$ when the sample size is large.
However, when the sample size is small, 
the estimated $\mathrm{QMI}(Z,Y)$ has high fluctuation.

\begin{figure*}[p]
  \begin{subfigure}[b]{0.50\linewidth}
  \centering
    \includegraphics[width=1\textwidth]{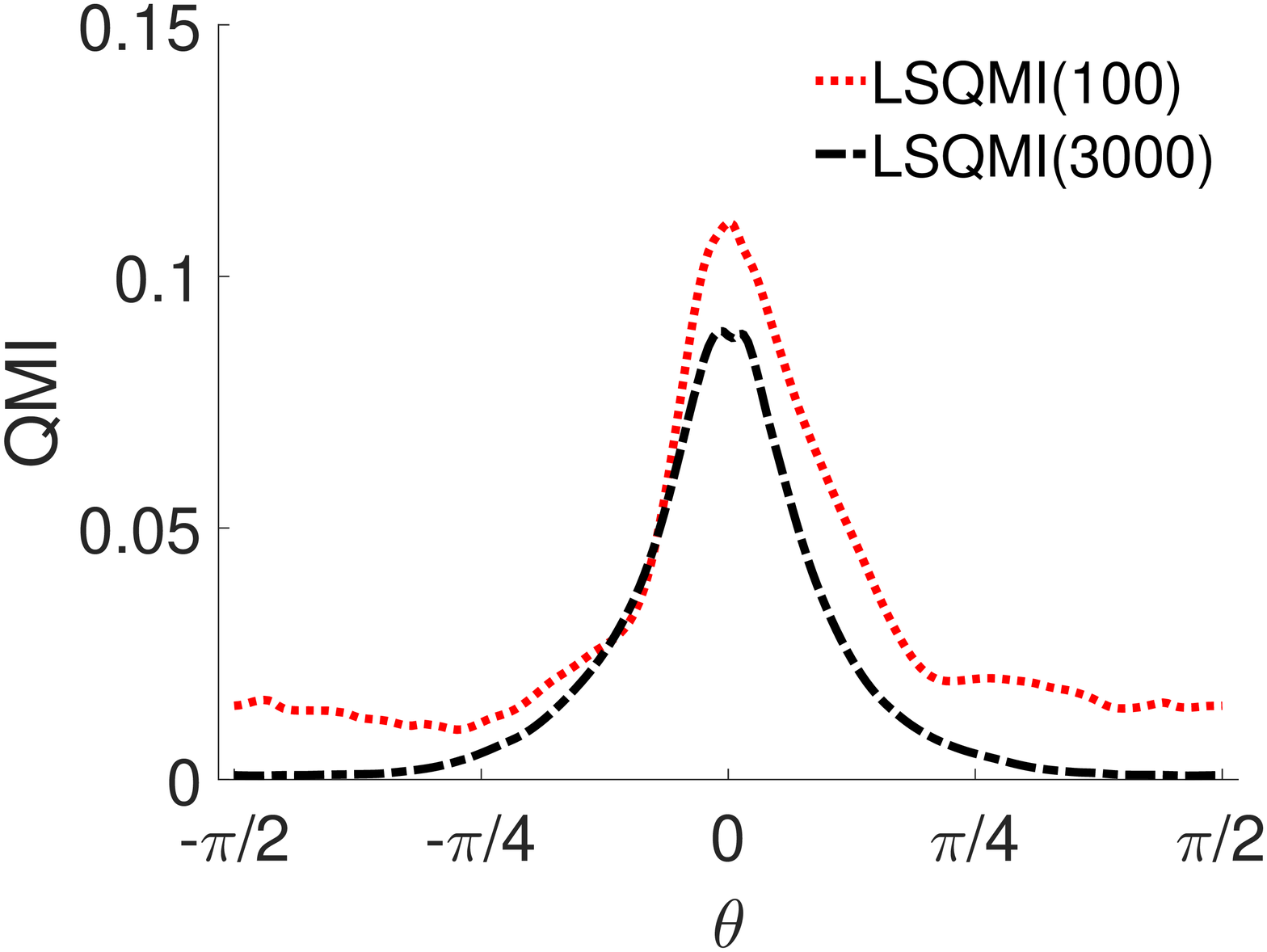}
    \includegraphics[width=1\textwidth]{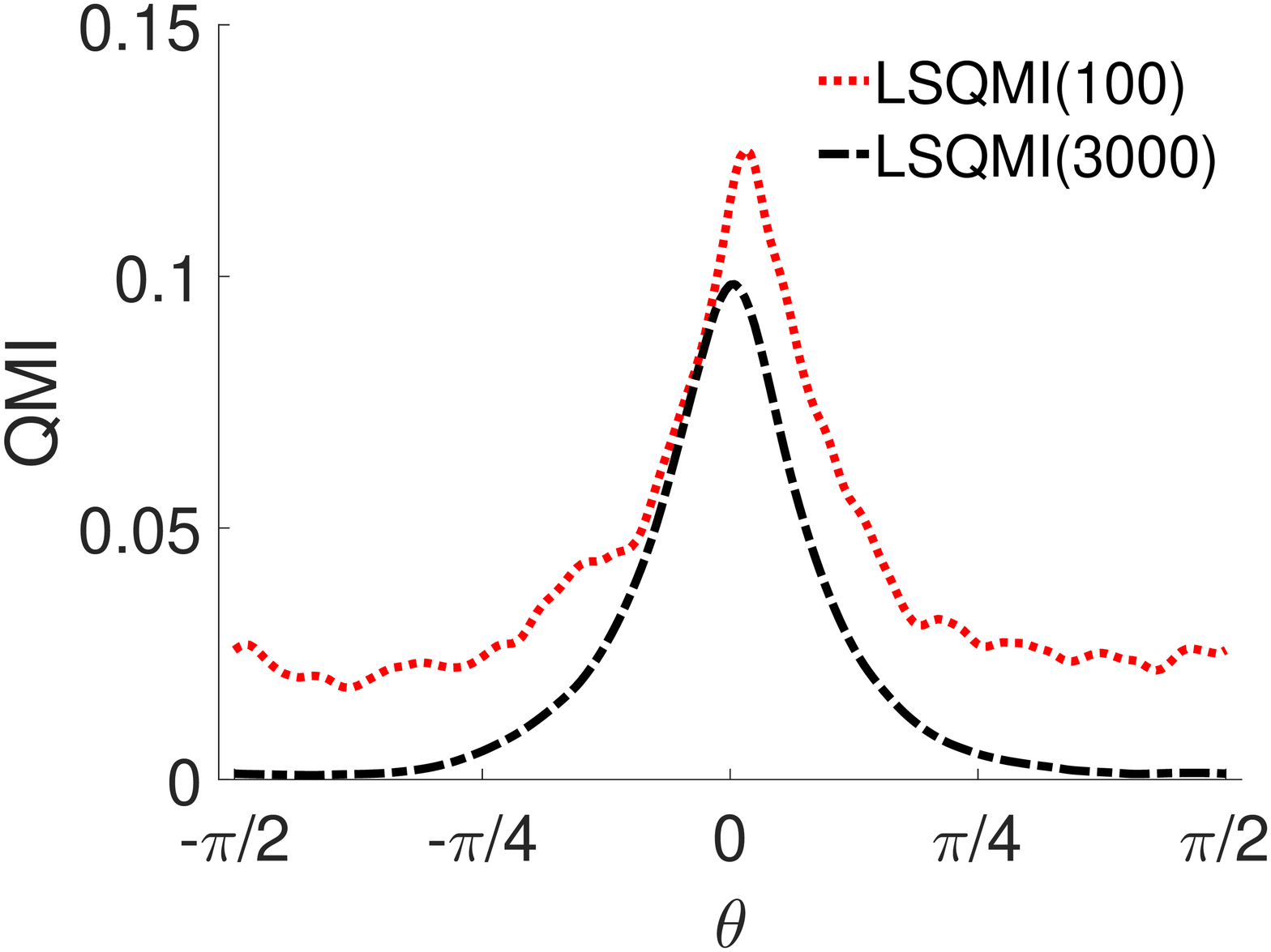}
    \includegraphics[width=1\textwidth]{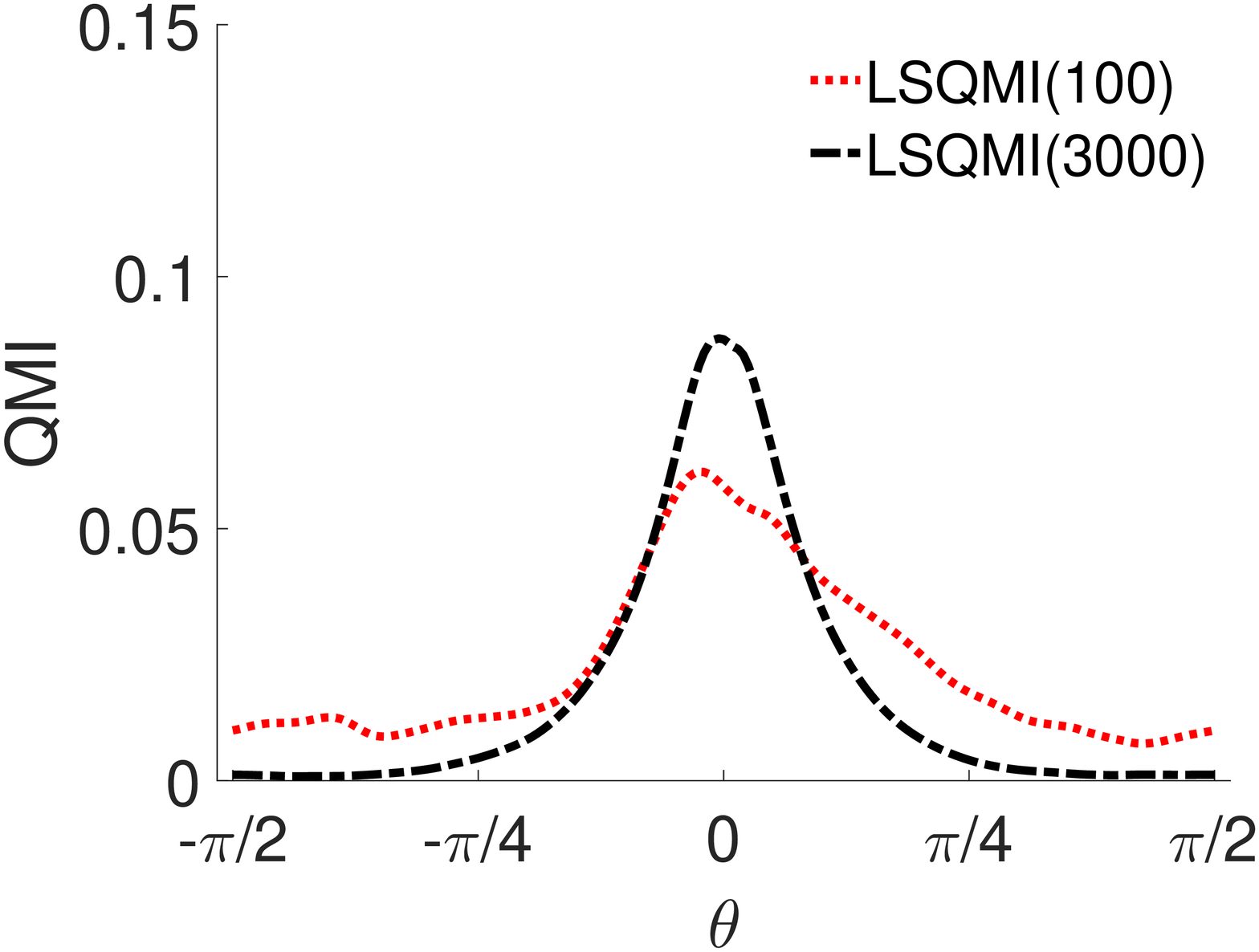}
    \includegraphics[width=1\textwidth]{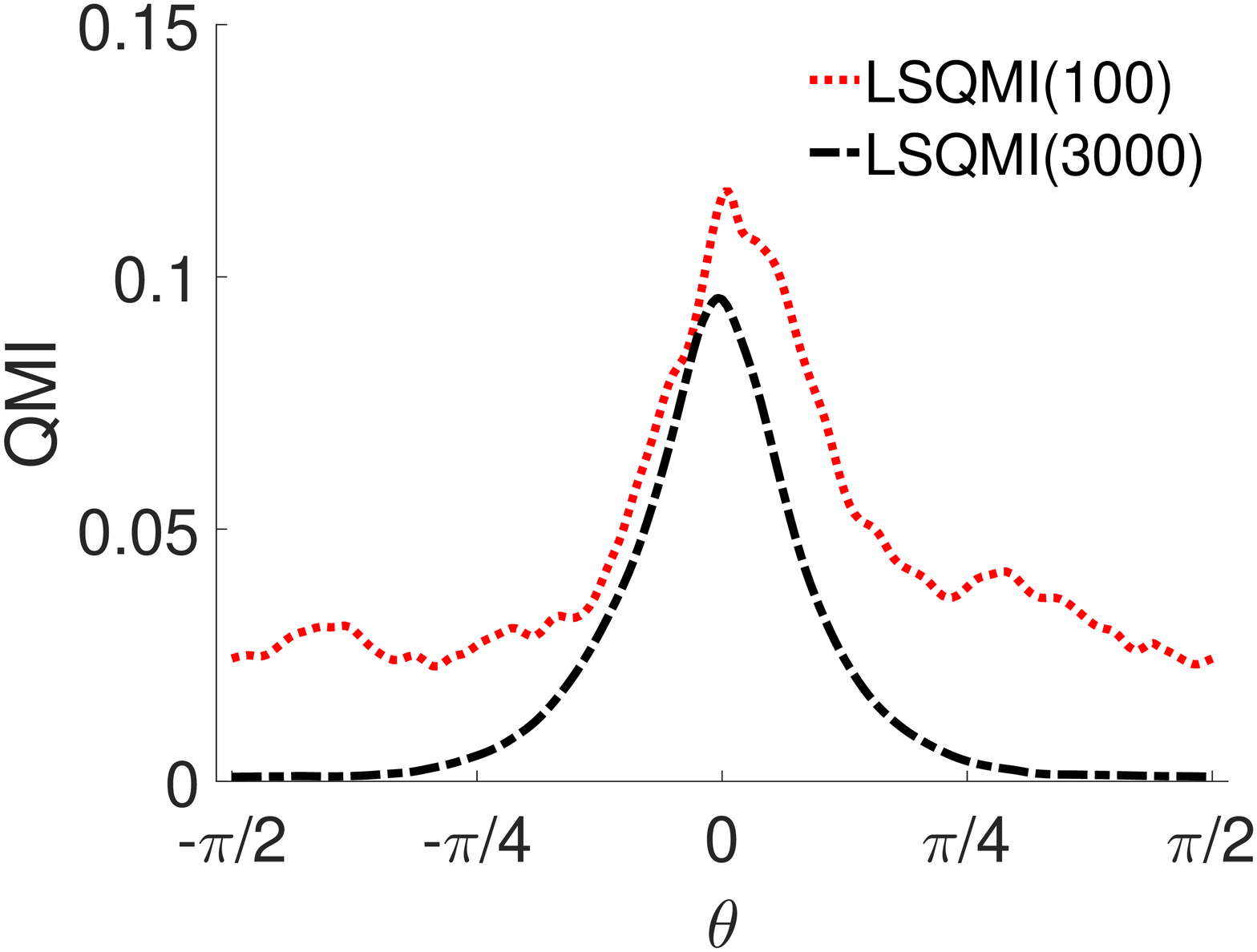}
    \subcaption{The estimated QMI.}
    \label{fig_qmi_illust}
  \end{subfigure}
  \begin{subfigure}[b]{0.50\linewidth}
  \centering
    \includegraphics[width=1\textwidth]{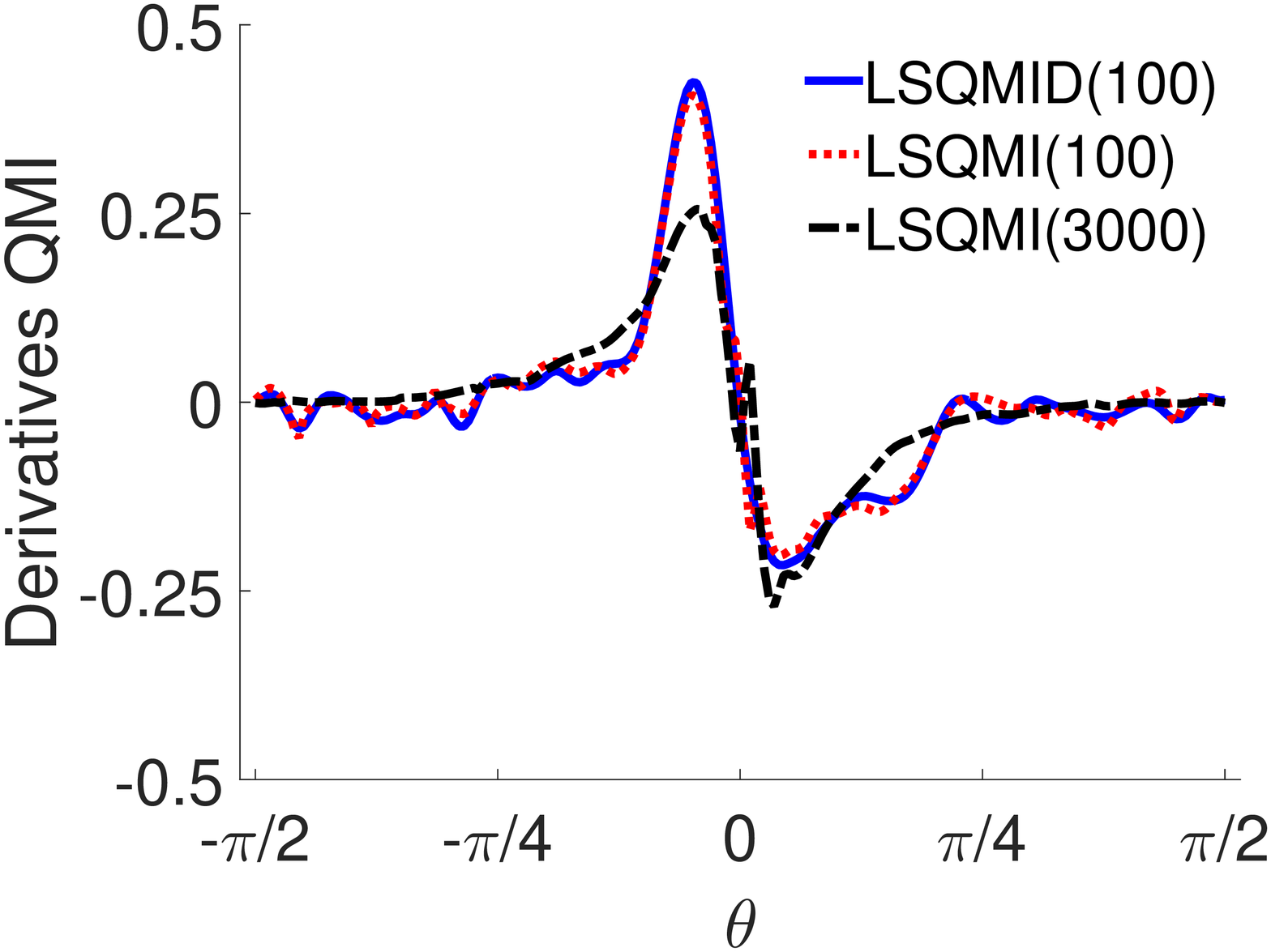}
    \includegraphics[width=1\textwidth]{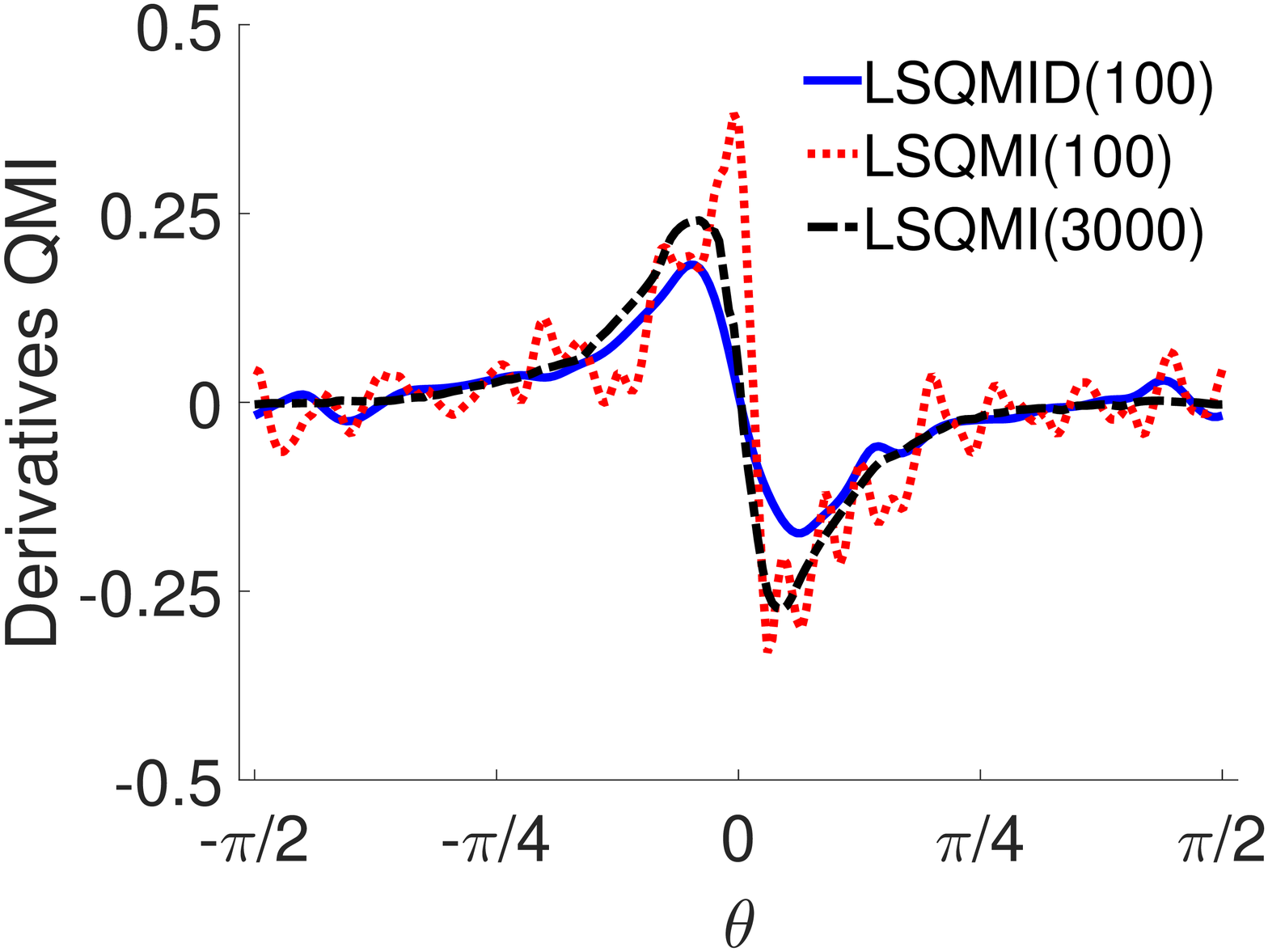}
    \includegraphics[width=1\textwidth]{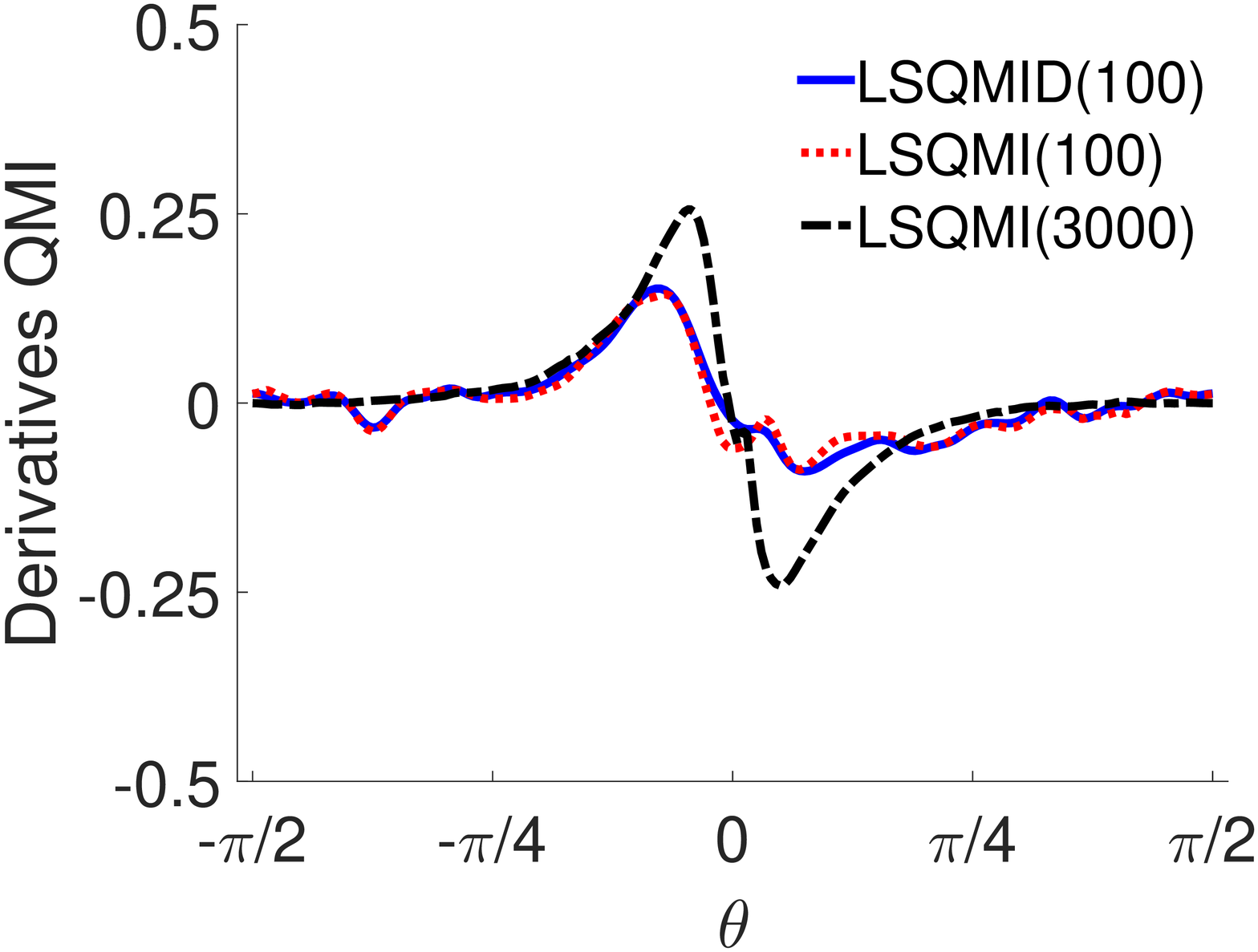}
    \includegraphics[width=1\textwidth]{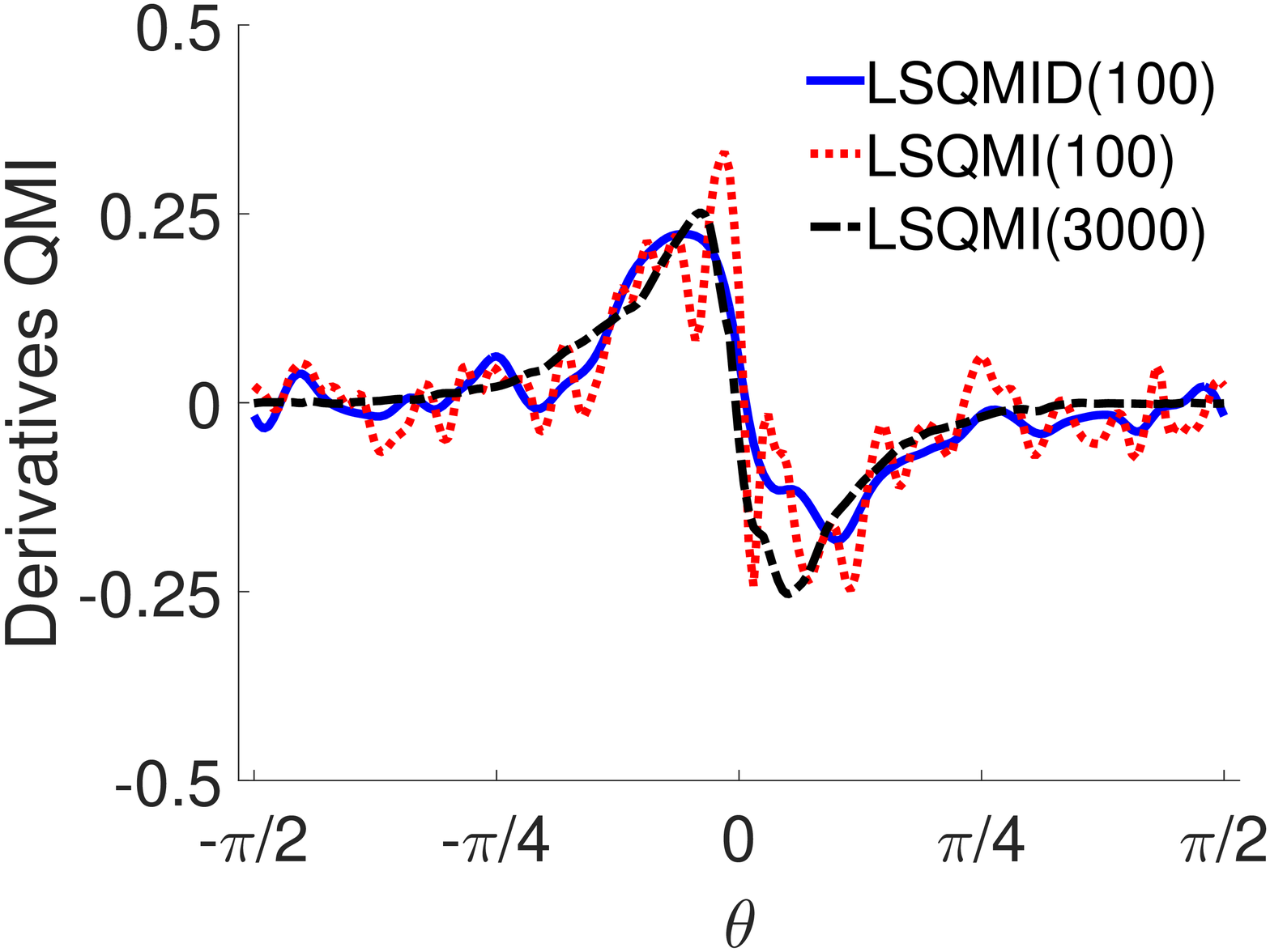}
    \subcaption{The estimated derivative of QMI.}
    \label{fig_dqmi_illust}
  \end{subfigure}
  \caption{Examples of the estimated QMI 
  and the estimated derivative of QMI.
  The left column shows the estimated $\mathrm{QMI}(Z,Y)$, and the right column shows the estimated derivative of $\mathrm{QMI}(Z,Y)$ w.r.t.~$\theta$.
  Each row indicates each experiment trial.}
  \label{fig_qmi_dqmi_illust}
\end{figure*}

Figure~\ref{fig_dqmi_illust_avg} shows
the averaged value over 20 experiment trials
of the derivative of $\mathrm{QMI}(Z,Y)$ w.r.t.~$\theta$ 
computed by LSQMI(3000), LSQMI(100), 
and the proposed method with $n=100$ 
which is denoted by LSQMID(100). 
For the proposed method, we perform cross validation at $\theta = 0$ 
and use the chosen tuning parameters for all values of $\theta$.
The result shows that LSQMID(100) gives a smoother estimate
than LSQMI(100) which has high fluctuation.
To further explain the cause of the fluctuation of LSQMI(100), 
we plot experiment results of 4 trials in Figure~\ref{fig_qmi_dqmi_illust}, where
the left column corresponds to the value of the estimated $\mathrm{QMI}(Z,Y)$
while the right column corresponds to the value of the estimated 
derivative of $\mathrm{QMI}(Z,Y)$ w.r.t.~$\theta$.
These results show that for LSQMI(100), 
a small fluctuation in the estimated QMI 
can cause a large fluctuation in the estimated derivative of QMI.
On the other hand, LSQMID 
directly estimates the derivative of QMI
and thus does not suffer from this problem.

\subsection{Artificial Datasets}
\begin{figure}[t]
  \begin{subfigure}[b]{0.50\linewidth}
  \centering
    \includegraphics[width=1\textwidth]{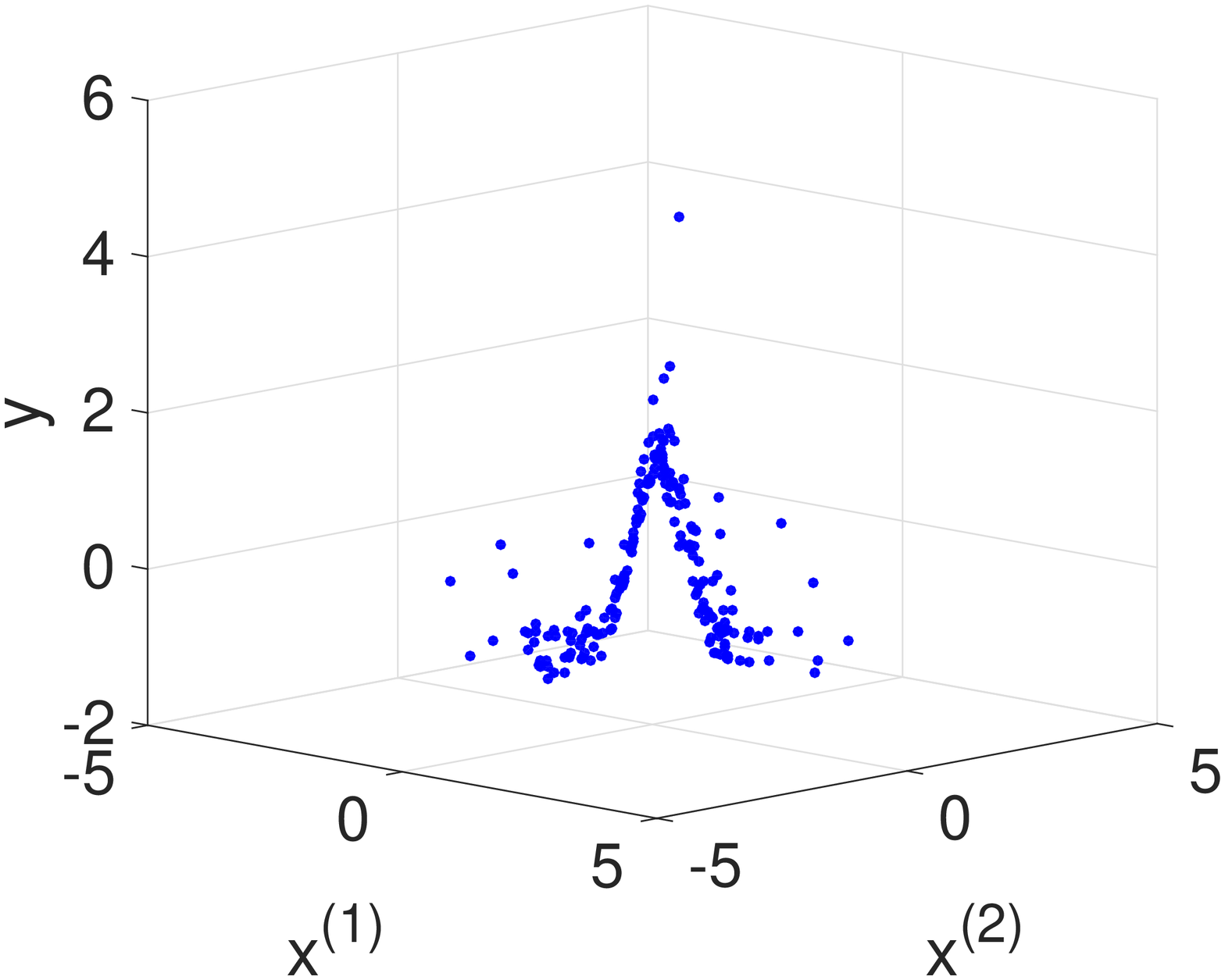}
    \subcaption{Dataset \textbf{A} with $n=200$}
  \end{subfigure}
  \begin{subfigure}[b]{0.50\linewidth}
  \centering
    \includegraphics[width=1\textwidth]{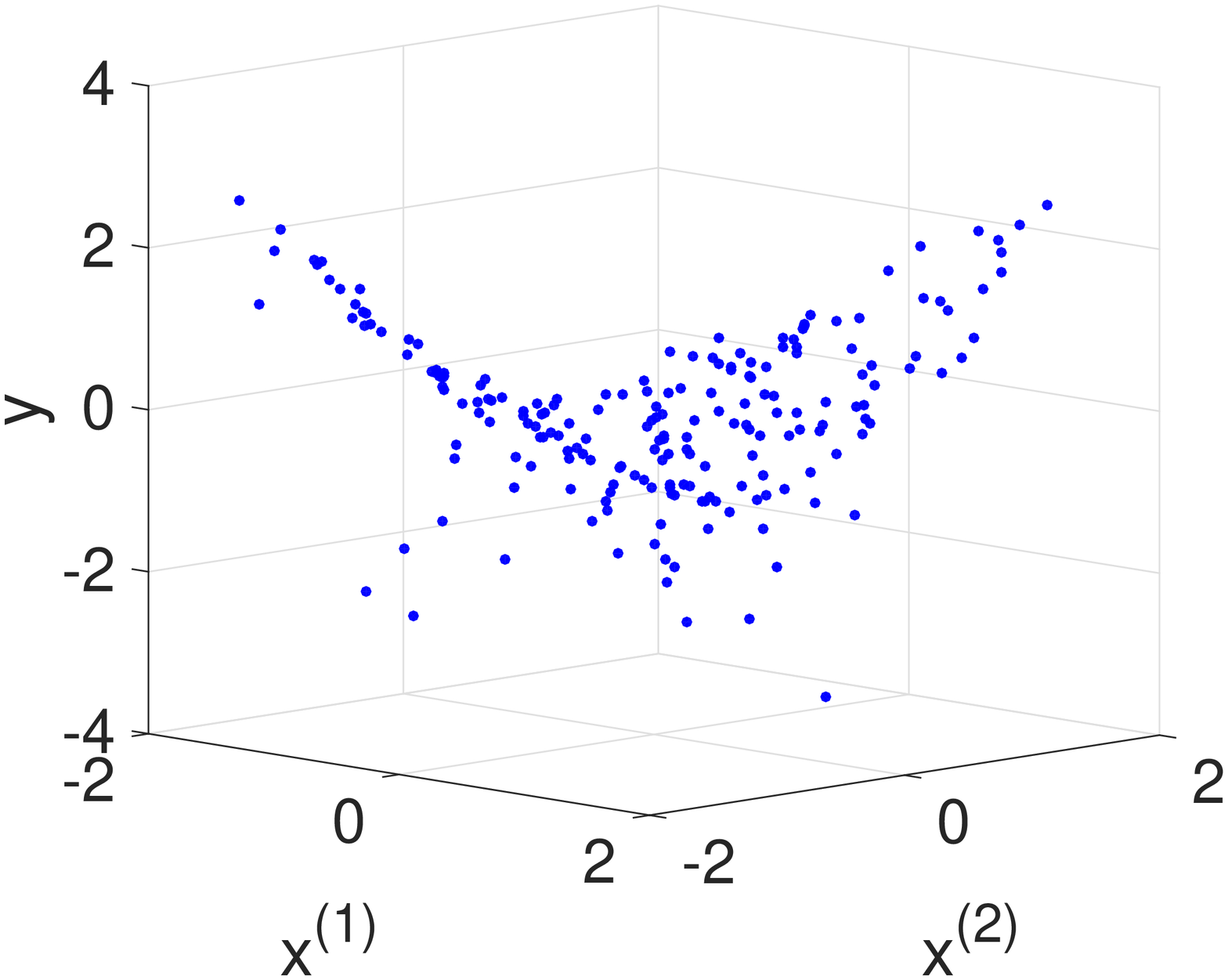}
    \subcaption{Dataset \textbf{B} with $n=200$}
  \end{subfigure}
  \begin{subfigure}[b]{0.50\linewidth}
  \centering
    \includegraphics[width=1\textwidth]{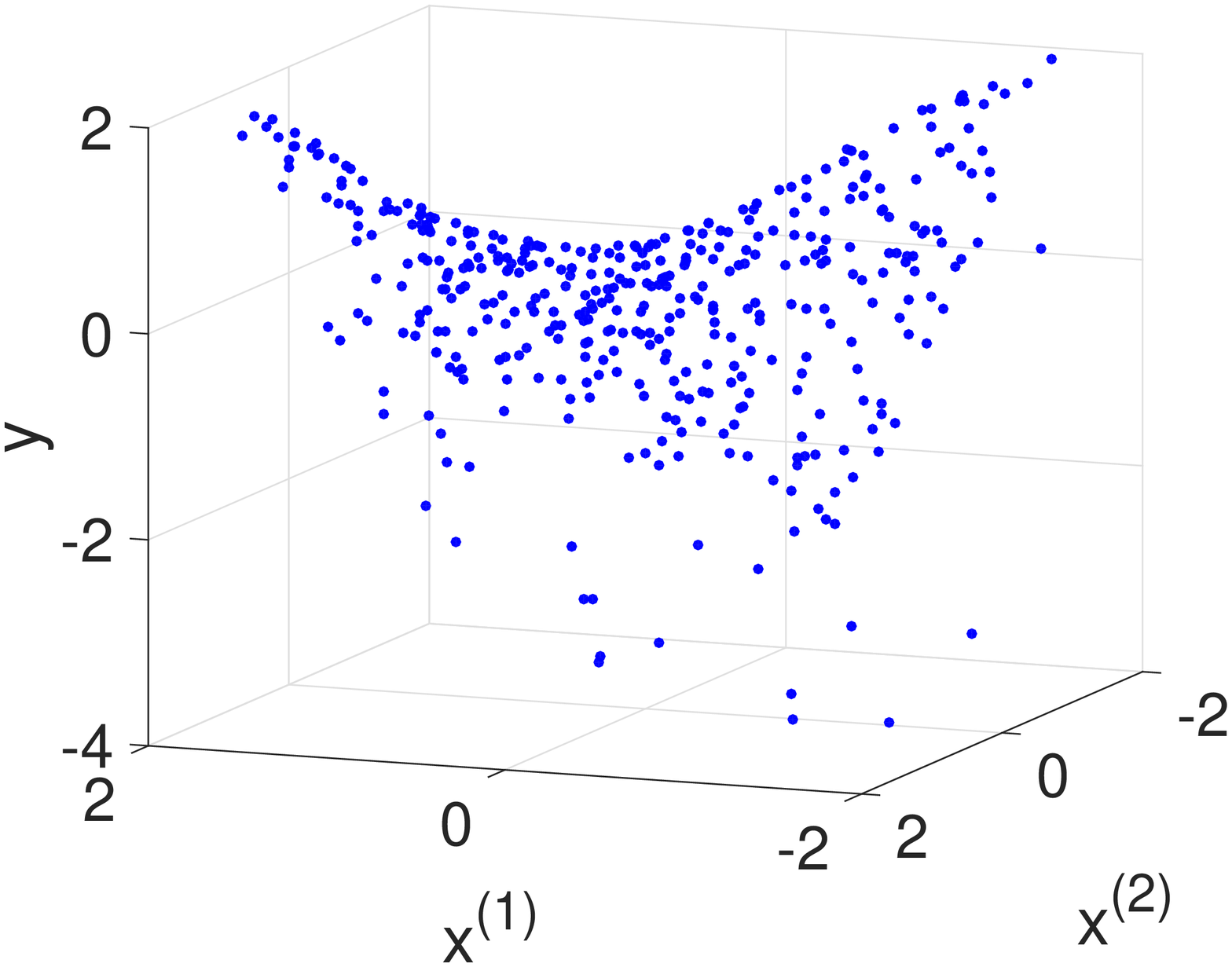}
    \subcaption{Dataset \textbf{C} with $n=400$}
  \end{subfigure}
  \begin{subfigure}[b]{0.50\linewidth}
  \centering
    \includegraphics[width=1\textwidth]{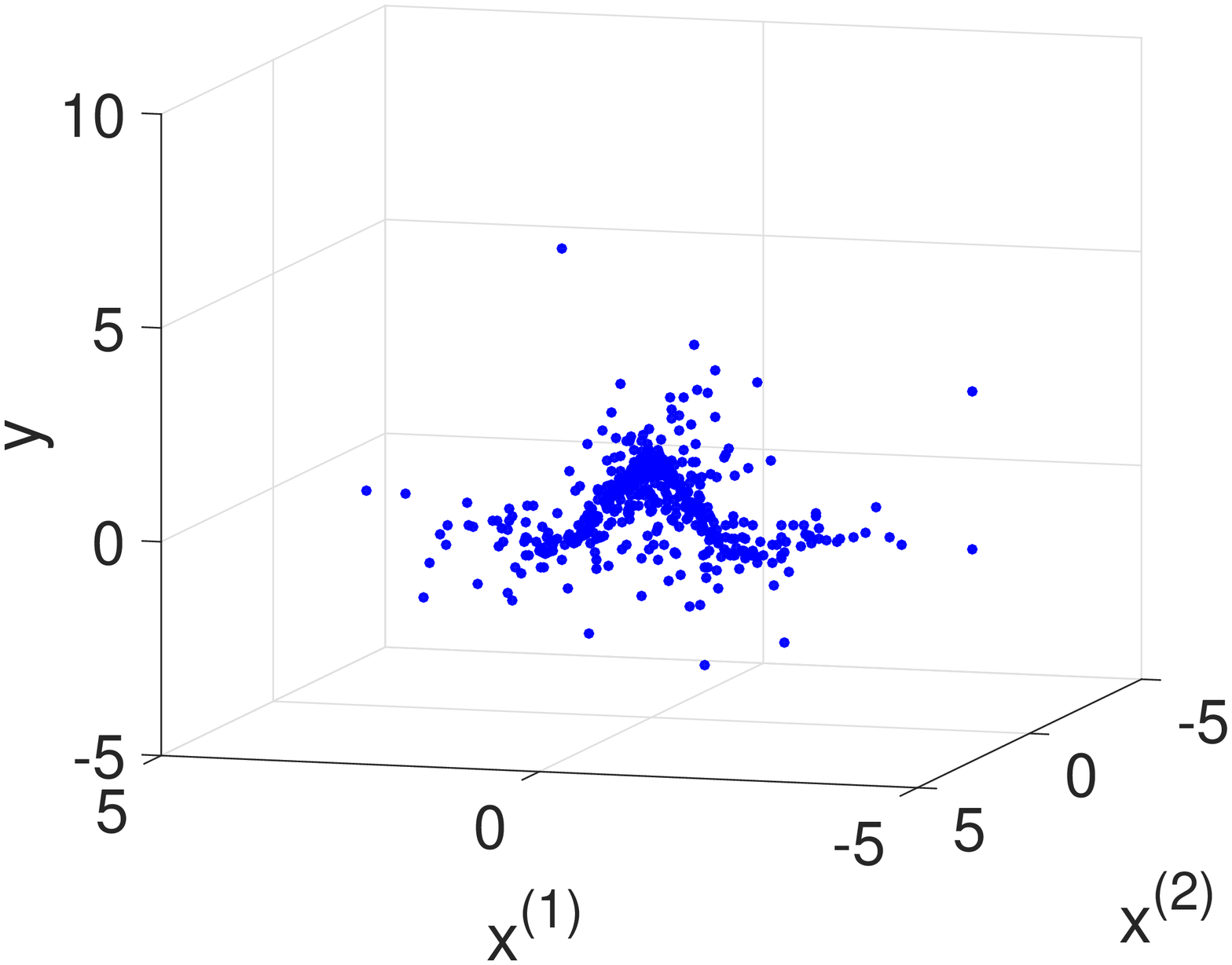}
    \subcaption{Dataset \textbf{D} with $n=500$}
  \end{subfigure}
  \caption{Artificial datasets.}
  \label{fig_art_data}
\end{figure}

Next, we evaluate the usefulness of the proposed method
in supervised dimension reduction using artificial datasets.
Firstly, let $\mathrm{U}(a,b)$ denote the 
uniform distribution over an interval $\left[a, b\right]$, 
$\Gamma(a,b)$ denote the gamma distribution with shape parameter $a$ and scale parameter $b$,
and $\mathrm{Laplace}(a,b)$ denote the Laplace distribution with mean $a$ and scale parameter $b$.
Then we consider the input $\bx$ with $d_{\mathrm{\boldsymbol{x}}} = 5$,
the output $y$ with $d_{\mathrm{\boldsymbol{y}}} = 1$,
and the optimal matrix $\bW_{\mathrm{opt}}$ (including their rotations) as follows:
\begin{description}
	\item[Dataset A:]
	For $\epsilon \sim \Gamma(0.25, 0.25)$, we use
	\begin{align*}
	\bx &\sim \mathcal{N}(\boldsymbol{0}_5,\boldsymbol{I}_5), \\
	y &= \exp(- \frac{(x^{(1)} + x^{(2)})^2}{0.5} ) + \epsilon,\\
	\bW_{\mathrm{opt}} &= \begin{bmatrix}
	\frac{1}{\sqrt{2}} & \frac{1}{\sqrt{2}} & 0 & 0 & 0
	\end{bmatrix}.
	\end{align*}
	
	\item[Dataset B:]
	For $\epsilon \sim \Gamma(0.25, 0.5)$ and $i \in \{1,\dots,5\}$, we use
	\begin{align*}
	x^{(i)} &\sim \mathrm{U}(-1,1), \\
	z &= \frac{1}{\sqrt{5}}(x^{(1)} + 2x^{(2)}), \\
	y &= z \sin(z) - \epsilon, \\
	\bW_{\mathrm{opt}} &= \begin{bmatrix}
	\frac{1}{\sqrt{5}} & \frac{2}{\sqrt{5}} & 0 & 0 & 0
	\end{bmatrix}.
	\end{align*}
	
	\item[Dataset C:]
	For $\epsilon \sim \Gamma(0.25, 0.5)$ and $i \in \{1,\dots,5\}$, we use
	\begin{align*}
	x^{(i)} &\sim \mathrm{U}(-1,1), \\
	y &= \frac{1}{\sqrt{2}}x^{(1)}x^{(2)} - \epsilon, \\
	\bW_{\mathrm{opt}} &= \begin{bmatrix}
	1 & 0 & 0 & 0 & 0 \\
	0 & 1 & 0 & 0 & 0
	\end{bmatrix}.
	\end{align*}
	
	\item[Dataset D:]
	For $\epsilon \sim \mathcal{N}(0, 0.25)$ and $i \in \{1,\dots,5\}$, we use
	\begin{align*}
	x^{(i)} &\sim \mathrm{Laplace}(0,0.5), \\
	y &= \mathrm{sinc}(\frac{x^{(1)}\pi}{2}) + x^{(2)}\epsilon, \\
	\bW_{\mathrm{opt}} &= \begin{bmatrix}
	1 & 0 & 0 & 0 & 0 \\
	0 & 1 & 0 & 0 & 0
	\end{bmatrix}.
	\end{align*}
\end{description}
For the datasets \textbf{A}, \textbf{B}, and \textbf{C},
$\epsilon$ is an additive gamma noise,
while for the datasets \textbf{D}, $\epsilon$ is a multiplicative Gaussian noise.
Figure~\ref{fig_art_data} shows the plot of these datasets (after standardization).
Note the presence of outliers in the datasets.

To estimate $\bW$ from $\{(\bx_i, y_i)\}_{i=1}^n$,
we execute the following methods:
\begin{description}
	\item[LSQMID:]
	The proposed method.
	Supervised dimension reduction is performed by maximizing $\mathrm{QMI}(Z,Y)$
	where the derivative of $\mathrm{QMI}(Z,Y)$ is estimated by the proposed method.
	The solution $\widehat{\bW}$ is obtained 
	by fixed-point iteration. 
	
	\item[LSQMI:]
	Supervised dimension reduction is performed by maximizing $\mathrm{QMI}(Z,Y)$
	where $\mathrm{QMI}(Z,Y)$ is estimated by LSQMI and 
	the derivative of $\mathrm{QMI}(Z,Y)$ w.r.t.~$\bW$ is computed from the QMI estimator.
	The solution $\widehat{\bW}$ is obtained by gradient ascent 
	with linesearch over the Grassmann manifold
	\footnote{We use the manifold optimization toolbox \citep{manopt} to perform the optimization.}. 
	
	\item[LSDR \textmd{\citep{DBLP:journals/neco/SuzukiS13}}:]
	Supervised dimension reduction is performed by maximizing $\mathrm{SMI}(Z,Y)$.
	The solution $\widehat{\bW}$ is obtained 
	by gradient ascent  with linesearch  over the Grassmann manifold 
	\footnote{We use the program code: \url{http://www.ms.k.u-tokyo.ac.jp/software.html#LSDR}}.
	
	\item[dMAVE \textmd{\citep{xia2007}}:] 
	Supervised dimension reduction is performed by minimizing an error of 
	the local linear smoother of the conditional density $p(y|\bz)$. 
	The solution $\widehat{\bW}$ is obtained by alternatively solving quadratic programming problems
	\footnote{We use the program code: \url{http://www.stat.nus.edu.sg/~staxyc/}}.
	
	\item[KDR \textmd{\citep{fukumizu2009}}:] 
	Supervised dimension reduction is performed by minimizing the trace of the
	conditional covariance operator $\Sigma_{\bY\bY|\bZ}$. 
	The solution $\widehat{\bW}$ is obtained by 
	gradient descent  with linesearch over the Stiefel manifold
	\footnote{We use the program code: \url{http://www.ism.ac.jp/~fukumizu/software.html}}.
\end{description}
For methods which require initial solutions, 
i.e., LSQMID, LSQMI, and LSDR, 
we randomly generate 10 orthonormal matrices 
and use them as the initial solutions.
For dMAVE and KDR, we use a solution obtained 
by dOPG and gKDR, respectively, as the initial solution.
Finally, the obtained solution $\widehat{\bW}$ 
is evaluated by the dimension reduction error defined as
\begin{align*}
\mathrm{Error}_{\mathrm{DR}} 
&= \| \bW_{\mathrm{opt}}^\top\bW_{\mathrm{opt}} 
    - \widehat{\bW}^\top\widehat{\bW}
\|_{\mathrm{Frobenius}},
\end{align*}
where $\| \cdot \|_{\mathrm{Frobenius}}$ denotes the Frobenius norm of a matrix.

Table~\ref{table_art} 
shows the mean and standard error over 30 experiment trials 
of the dimension reduction error on the 
artificial datasets with different sample size.
The results show that the proposed method works well overall.
LSDR performs well especially for dataset \textbf{A} and \textbf{B}.
KDR also performs well overall.
However, its performance is quite unstable for dataset \textbf{B},
which can be seen by relatively large standard errors.
This is because gKDR might provide 
a poor initial solution to KDR in some experiment trials,
which makes KDR fails to find a good solution.

On the other hand, both LSQMI and dMAVE do not perform well overall.
LSQMI tends to be unstable and works very poorly 
especially when the sample size is small, except for dataset \textbf{D}.
The cause of this failure could be 
the high fluctuation of the derivative of QMI by LSQMI,
as shown previously in the illustrative experiment.
Although the solution of dMAVE is quite stable,
its performance is not overall comparable to the other methods.
This is because the model selection strategy 
in dMAVE did not perform well for these datasets.
\begin{sidewaystable}
\small
\centering
\caption{Mean and standard error of the dimension reduction error over 30 trials for artificial datasets. The best method in term of the mean error and comparable methods according to the paired \textit{t-test} at the significance level $5\%$ are specified by bold face.
}
\vspace{1mm}
\begin{tabular}{c | c | c | c | c | c | c }   
\hline
Dataset & $n$ & LSQMID &  LSQMI & LSDR & dMAVE & KDR  \\ \hline
{\multirow{2}{*}{\textbf{A}}} & 100
		 & $\boldsymbol{0.126(0.039)}$
		 & $0.394(0.094)$
		 & $\boldsymbol{0.073(0.007)}$
		 & $0.123(0.009)$
		 & $\boldsymbol{0.077(0.008)}$ \\ 
{} & 200
		 & $\boldsymbol{0.046(0.006)}$
		 & $\boldsymbol{0.077(0.035)}$
		 & $\boldsymbol{0.046(0.005)}$
		 & $0.082(0.007)$
		 & $\boldsymbol{0.048(0.006)}$ \\ \hline
{\multirow{2}{*}{\textbf{B}}} & 100 
		 & $\boldsymbol{0.079(0.009)}$
		 & $0.488(0.097)$
		 & $\boldsymbol{0.072(0.006)}$
		 & $0.128(0.010)$
		 & $0.306(0.088)$ \\ 
{} & 200
		 & $\boldsymbol{0.041(0.004)}$
		 & $0.174(0.064)$
		 & $\boldsymbol{0.038(0.003)}$
		 & $0.078(0.006)$
		 & $\boldsymbol{0.099(0.045)}$ \\ \hline
{\multirow{2}{*}{\textbf{C}}} & 200 
		 & $\boldsymbol{0.192(0.026)}$
		 & $0.633(0.100)$
		 & $0.203(0.011)$
		 & $\boldsymbol{0.146(0.010)}$
		 & $\boldsymbol{0.155(0.010)}$ \\ 
{} & 400 
		 & $\boldsymbol{0.084(0.005)}$
		 & $0.108(0.011)$
		 & $0.128(0.006)$
		 & $0.105(0.006)$
		 & $\boldsymbol{0.090(0.007)}$ \\ \hline
{\multirow{2}{*}{\textbf{D}}} & 300   
		 & $\boldsymbol{0.245(0.050)}$
		 & $\boldsymbol{0.301(0.054)}$
		 & $\boldsymbol{0.286(0.032)}$
		 & $0.370(0.046)$
		 & $\boldsymbol{0.257(0.025)}$ \\ 
{} & 500 
		 & $\boldsymbol{0.128(0.017)}$
		 & $\boldsymbol{0.129(0.013)}$
		 & $0.185(0.015)$
		 & $0.263(0.038)$
		 & $0.198(0.013)$ \\ \hline
\end{tabular}
\label{table_art}
\vspace{4mm}
\centering
\caption{Mean and standard error of the root mean squared error over 30 trials for benchmark datasets. The best method in term of the mean error and comparable methods according to the paired \textit{t-test} at the significance level $5\%$ are specified by bold face.
}
\vspace{2mm}
\begin{tabular}{c | c |  c | c | c | c | c | c | c }   
\hline
Dataset & $n_{\mathrm{tr}}$ & $d_{\boldsymbol{\mathrm{\widetilde{x}}}}$  
& $d_{\boldsymbol{\mathrm{z}}}$ & 
LSQMID & LSQMI & LSDR & dMAVE & KDR \\ \hline
{\multirow{4}{*}{Fertility}} & {\multirow{4}{*}{50}} & {\multirow{4}{*}{14}} & 1
		 & $1.215(0.049)$
		 & $\boldsymbol{1.092(0.043)}$
		 & $1.315(0.043)$
		 & $1.321(0.063)$
		 & $\boldsymbol{1.116(0.050)}$ \\ 
{} & {} & {} & 2
		 & $\boldsymbol{1.051(0.045)}$
		 & $\boldsymbol{1.029(0.043)}$
		 & $1.199(0.031)$
		 & $1.340(0.052)$
		 & $1.104(0.044)$ \\ 
{} & {} & {} & 3
		 & $\boldsymbol{1.052(0.044)}$
		 & $\boldsymbol{1.038(0.047)}$
		 & $1.104(0.044)$
		 & $1.288(0.048)$
		 & $1.121(0.043)$ \\ 
{} & {} & {} & 4
		 & $\boldsymbol{1.046(0.042)}$
		 & $\boldsymbol{1.026(0.042)}$
		 & $1.092(0.039)$
		 & $1.271(0.033)$
		 & $1.146(0.044)$ \\ \hline
{\multirow{4}{*}{Yacht}} & {\multirow{4}{*}{100}} & {\multirow{4}{*}{11}} & 1
		 & $\boldsymbol{0.120(0.005)}$
		 & $0.546(0.042)$
		 & $0.180(0.012)$
		 & $0.213(0.017)$
		 & $\boldsymbol{0.124(0.007)}$ \\ 
{} & {} & {} & 2
		 & $\boldsymbol{0.154(0.011)}$
		 & $0.675(0.047)$
		 & $0.344(0.023)$
		 & $0.224(0.014)$
		 & $0.278(0.033)$ \\ 
{} & {} & {} & 3
		 & $\boldsymbol{0.314(0.024)}$
		 & $0.690(0.037)$
		 & $0.425(0.018)$
		 & $\boldsymbol{0.265(0.013)}$
		 & $0.353(0.028)$ \\ 
{} & {} & {} & 4
		 & $0.413(0.021)$
		 & $0.732(0.043)$
		 & $0.494(0.015)$
		 & $\boldsymbol{0.352(0.017)}$
		 & $0.399(0.012)$ \\  \hline
{\multirow{4}{*}{Concrete}} & {\multirow{4}{*}{200}} & {\multirow{4}{*}{13}} & 1
		 & $0.621(0.013)$
		 & $\boldsymbol{0.606(0.014)}$
		 & $0.606(0.008)$
		 & $\boldsymbol{0.582(0.006)}$
		 & $0.791(0.030)$ \\ 
{} & {} & {} & 2
		 & $0.568(0.010)$
		 & $0.591(0.009)$
		 & $0.568(0.010)$
		 & $\boldsymbol{0.529(0.009)}$
		 & $0.614(0.025)$ \\ 
{} & {} & {} & 3
		 & $\boldsymbol{0.557(0.009)}$
		 & $0.579(0.011)$
		 & $0.576(0.012)$
		 & $\boldsymbol{0.539(0.007)}$
		 & $0.579(0.016)$ \\ 
{} & {} & {} & 4
		 & $\boldsymbol{0.545(0.012)}$
		 & $0.667(0.025)$
		 & $0.568(0.010)$
		 & $\boldsymbol{0.540(0.008)}$
		 & $0.571(0.014)$ \\  \hline
{\multirow{4}{*}{Breast-cancer}} & {\multirow{4}{*}{200}} & {\multirow{4}{*}{15}} & 1
		 & $0.447(0.011)$
		 & $0.523(0.018)$
		 & $0.442(0.010)$
		 & $\boldsymbol{0.375(0.007)}$
		 & $0.447(0.012)$ \\
{} & {} & {} & 2
		 & $\boldsymbol{0.435(0.010)}$
		 & $0.473(0.012)$
		 & $\boldsymbol{0.437(0.009)}$
		 & $\boldsymbol{0.420(0.012)}$
		 & $0.454(0.014)$ \\  
{} & {} & {} & 3
		 & $\boldsymbol{0.376(0.004)}$
		 & $0.462(0.010)$
		 & $0.431(0.007)$
		 & $0.426(0.008)$
		 & $0.430(0.007)$ \\
{} & {} & {} & 4
		 & $\boldsymbol{0.377(0.005)}$
		 & $0.419(0.008)$
		 & $0.436(0.007)$
		 & $0.426(0.011)$
		 & $0.433(0.007)$ \\  \hline
{\multirow{4}{*}{Bike}} & {\multirow{4}{*}{300}} & {\multirow{4}{*}{19}} & 1
		 & $0.043(0.011)$
		 & $0.070(0.019)$
		 & $\boldsymbol{0.016(0.001)}$
		 & $0.139(0.051)$
		 & $0.513(0.059)$ \\
{} & {} & {} & 2
		 & $\boldsymbol{0.036(0.005)}$
		 & $\boldsymbol{0.035(0.003)}$
		 & $0.049(0.002)$
		 & $0.081(0.007)$
		 & $0.291(0.050)$ \\  
{} & {} & {} & 3
		 & $\boldsymbol{0.037(0.005)}$
		 & $\boldsymbol{0.032(0.003)}$
		 & $0.065(0.002)$
		 & $0.086(0.008)$
		 & $0.243(0.037)$ \\
{} & {} & {} & 4
		 & $\boldsymbol{0.060(0.006)}$
		 & $\boldsymbol{0.051(0.007)}$
		 & $0.077(0.002)$
		 & $0.071(0.005)$
		 & $0.213(0.029)$ \\  \hline 
\end{tabular}
\label{table_bench}
\end{sidewaystable}

\subsection{Benchmark Datasets}
Finally, we evaluate the proposed method in supervised dimension reduction
on UCI benchmark datasets \citep{Bache+Lichman:2013}.
For all datasets, we append the original input $\bx$ 
with noise features of dimensionality 5.
More specifically, for the original input $\bx$ 
with dimensionality $d_{\boldsymbol{\mathrm{x}}}$,
we consider the augmented input $\widetilde{\bx}$ 
with dimensionality 
$d_{\boldsymbol{\mathrm{\widetilde{x}}}} = 
d_{\boldsymbol{\mathrm{x}}} + 5$ as
\begin{align*}
\widetilde{\bx} = 
\begin{bmatrix}
\bx^\top, \gamma_1, \gamma_2, \gamma_3, \gamma_4, \gamma_5
\end{bmatrix}^\top,
\end{align*}
where $\gamma_i \sim \Gamma(1,2)$ 
for $i \in \{1, \dots, 5\}$.
Then we use the paired data 
$\{(\widetilde{\bx}_i, y_i )\}_{i=1}^n$
to perform experiments.
We randomly choose $n_{\mathrm{tr}}$ samples for training purposes,
and use the rest $n_{\mathrm{te}} = n - n_{\mathrm{tr}}$ for testing purposes.
We execute the supervised dimension reduction methods with target dimensionality 
$d_{\boldsymbol{\mathrm{z}}} \in \{1,2,3,4 \}$ to 
obtain solutions $\widehat{\bW}$.
Then we train a kernel ridge regressor~
$\widehat{y} = f(\widehat{\bW}\widetilde{\bx})$~
with the Gaussian kernel where the tuning parameters are chosen by 5-fold cross-validation.
Finally, we evaluate the regressor by the \textit{root mean squared error} (RMSE):
\begin{align*}
\mathrm{RMSE} = \sqrt{ \frac{1}{n_{\mathrm{te}}}\sum_{i=1}^{n_{\mathrm{te}}} 
   \left( y_i - f(\widehat{\bW}\widetilde{\bx}_i) \right)^2  }.
\end{align*}

Table~\ref{table_bench} shows the RMSE averaged 
over 30 trials for the benchmark experiments.
It shows that the proposed method performs well overall on all datasets.
LSQMI performs very well for the `Fertility' and `Bike' datasets,
but its performance is quite poor for the other datasets.
In contrast, dMAVE performs very well 
especially for the `Concrete' dataset
where it gives the best solutions for all value of $d_{\boldsymbol{\mathrm{z}}}$.
However, its performance is quite poor for the `Fertility' and `Bike' datasets.
Both LSDR and KDR do not perform well on these datasets.

\section{Conclusion}
\label{section:conclusion}

We proposed a novel supervised dimension reduction method 
based on efficient maximization of 
quadratic mutual information (QMI).
Our key idea was to \textit{directly} estimate the derivative of QMI 
without estimating QMI itself.
We firstly developed a method to directly estimate the derivative of QMI,
and then developed fixed-point iteration which 
efficiently uses the derivative estimator 
to find a maximizer of QMI.
In addition to the robustness against outliers thanks to the property of QMI, 
the proposed method is widely applicable 
because it does not require any assumption on the data distribution
and tuning parameters can 
be objectively chosen via cross-validation.
The experiment results on artificial and benchmark datasets 
showed that the proposed method is promising.



\end{document}